\documentclass[numbers]{article}

\usepackage[final]{neurips_2021}

\usepackage[utf8]{inputenc} % allow utf-8 input
\usepackage[T1]{fontenc}    % use 8-bit T1 fonts
\usepackage{hyperref}       % hyperlinks
\usepackage{url}            % simple URL typesetting
\usepackage{booktabs}       % professional-quality tables
\usepackage{amsfonts}       % blackboard math symbols
\usepackage{nicefrac}       % compact symbols for 1/2, etc.
\usepackage{microtype}      % microtypography
\usepackage{xcolor}         % colors

\usepackage{tikz} % nice language for creating drawings
\usepackage{float}
\usepackage{amsmath,amsfonts,bm}

\def\eqref#1{equation~\ref{#1}}
\def\Eqref#1{Equation~\ref{#1}}
\def\1{\bm{1}}

\DeclareMathAlphabet{\mathsfit}{\encodingdefault}{\sfdefault}{m}{sl}
\SetMathAlphabet{\mathsfit}{bold}{\encodingdefault}{\sfdefault}{bx}{n}

\newcommand{\E}{\mathbb{E}}

\newcommand{\KL}{D_{\mathrm{KL}}}

\usepackage{afterpage}
\usepackage{pdflscape}
\usepackage{caption}
\usepackage{graphicx}
\usepackage{subfigure}
\usepackage{sidecap}
\usepackage{wrapfig}

\newcommand{\newtext}[1]{\textcolor{black} {#1}}

\title{Deep Variational Semi-Supervised Novelty Detection}

\author{%
  Tal Daniel\\
  ECE Department, Technion\\
  Haifa, Israel \\
  \texttt{taldanielm@campus.technion.ac.il} \\
   \And
   Thanard Kurutach \\
   EECS Department, UC Berkeley \\
   California, USA \\
   \AND
   Aviv Tamar \\
   ECE Department, Technion \\
   Haifa, Israel \\
}

\begin{document}

\maketitle

\begin{abstract}
In anomaly detection (AD), one seeks to identify whether a test sample is abnormal, given a data set of normal samples. A recent and promising approach to AD relies on deep generative models, such as variational autoencoders (VAEs), for unsupervised learning of the normal data distribution. In semi-supervised AD (SSAD), the data also includes a small sample of labeled anomalies. In this work, we propose two variational methods for training VAEs for SSAD. The intuitive idea in both methods is to train the encoder to `separate' between latent vectors for normal and outlier data. We show that this idea can be derived from principled probabilistic formulations of the problem, and  propose simple and effective algorithms. Our methods can be applied to various data types, as we demonstrate on SSAD datasets ranging from natural images to astronomy and medicine, can be combined with any VAE model architecture, \newtext{and are naturally compatible with ensembling}. When comparing to state-of-the-art SSAD methods that are not specific to particular data types, we obtain marked improvement in outlier detection. 
\end{abstract}

\section{Introduction}
\label{intro}
% AD is important
Anomaly detection (AD) -- the task of identifying abnormal samples with respect to some normal data -- has applications in domains ranging from health-care, to security, and robotics~\citep{pimentel2014review}. In its common formulation, training data is provided only for normal samples, while at test time, anomalous samples need to be detected. In the probabilistic AD approach, a model of the normal data distribution is learned, and the likelihood of a test sample under this model is thresholded for classification as normal or not. 
% \TD{\cite{nalisnick2018do} performed an extensive work on deep generative models for out-of-distribution detection and concluded that they are unreliable.} 
Recently, deep generative models such as variational autoencoders (VAEs, \cite{kingma2013auto}) and generative adversarial networks~\citep{goodfellow2014generative} have shown promise for learning data distributions in AD~\citep{an2015variational,suh2016echo,schlegl2017unsupervised,wang2017safer, Zisselman_2020_CVPR}. 

\begin{figure}[t]
% \begin{SCfigure}
% \begin{figure}
\centering
\includegraphics[width=0.8\linewidth, trim={5cm 1cm 5cm 1cm},clip]{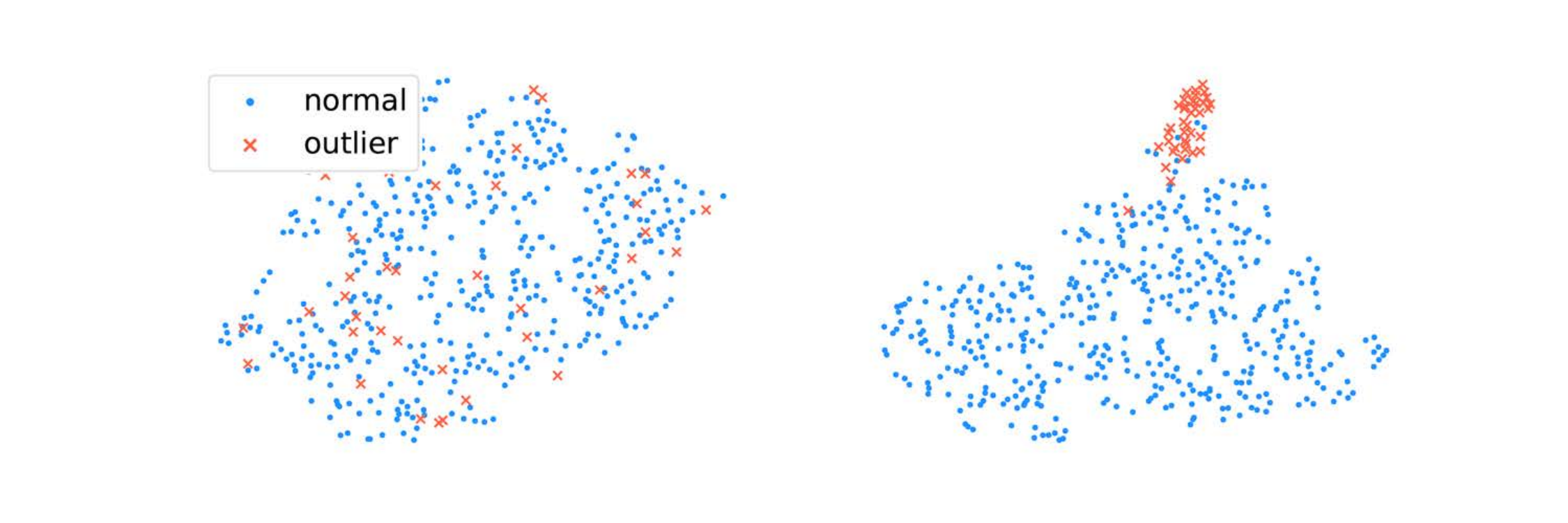}
\caption{t-SNE of the latent space of a conventional VAE (left), and our proposed dual-prior VAE (right), trained on Thyroid dataset, and evaluated on an unseen test sample. Latent vectors for outliers (red crosses) in dual-prior VAE are pushed away from normal samples.}
\label{fig:tsne_latent}
% \end{figure}
% \end{SCfigure}
\end{figure}

% SSAD is important. Deep SSAD is important.
Here, we consider the setting of \textit{semi-supervised} AD (SSAD), where in addition to the normal samples, a small sample of labeled anomalies is provided~\citep{gornitz2013toward}. Most importantly, \textit{this set is too small to represent the range of possible anomalies}. For example, consider normal data representing the digit `0' in the MNIST dataset, and being given a small sample from the digit `1'as anomalies. The goal in SSAD is to use this data to better classify as anomalies \textit{all the digits} `1'-`9'. In this case, it is clear that classification methods (either supervised or semi-supervised) are not suitable.
Instead, most approaches are based on `fixing' an unsupervised AD method to correctly classify the labeled anomalies, while still maintaining AD capabilities for unseen outliers~\citep{gornitz2013toward,munoz2010semisupervised,ruff2019}.  
% \TD{Our work show that in the unsupervised setting, the results align with the conclusion in \cite{nalisnick2018do}. However, when incorporating labeled samples in a semi-supervised setting, there is a substantial improvement.}

We present a variational approach for learning data distributions in the SSAD problem setting. We base our method on the VAE, and modify the training objective to account for the labeled outlier data. We propose two formulations for this problem. The first maximizes the log-likelihood of normal samples, while minimizing the log-likelihood of outliers, and we effectively optimize this objective by combining the standard evidence lower bound (ELBO) with the $\chi$ upper bound (CUBO, \cite{dieng2017variational}). The second method is based on separating the VAE prior between normal and outlier samples. Effectively, both methods have a similar intuitive interpretation: they modify the VAE encoder to push outlier samples away from the prior distribution (Fig.~\ref{fig:tsne_latent}). 

The state-of-the-art Deep-SAD SSAD method~\citep{ruff2019} also separates the encoding of normal and outlier samples. In comparison, by building on variational inference theory, our approach does not place any restriction on the network architecture as~\citet{ruff2019}, and can be used to modify any VAE to account for outliers. Additionally, our derivation allows a principled use of ensembles, which we show to improve performance significantly.

% Brief summary of results.
We evaluate our methods in the comprehensive SSAD test-suite of \citet{ruff2019}, which includes both image data and low-dimensional data sets from astronomy, medicine, and other domains. Our method is on par with or better than the state-of-the-art.
% Deep-SAD, and report a marked improvement in performance, \TD{or on par with the current SOTA}, compared to both shallow and deep methods. 
For some image data sets, \citet{nalisnick2018do} argued that VAEs are not informative novelty detectors. However, as demonstrated by~\citet{daniel2021soft}, incorporating a likelihood-based adversarial term in the objective of VAE results in excellent out-of-distribution detection in the same setting as \citet{nalisnick2018do}. As we show here, even $1$\% of labeled anomalies can significantly improve detection in the setting of~\citet{ruff2019}.

Additionally, to the best of our knowledge, our approaches are the first principled probabilistic formulation for training VAEs with negative examples.
We demonstrate this idea, and further establish the flexibility of our method, by modifying a conditional VAE used for generating sampling distributions for robotic motion planning to not generate way points that collide with obstacles~\citep{ichter_mp_18}, and also, on learning to identify novel sentiments in text, based on a VAE for sentences~\citep{bowman-etal-2016-generating}.

\section{Related Work}
\label{rel_work}
Anomaly detection, a.k.a.~outlier detection or novelty detection, is an active field of research. Shallow methods include one-class SVM (OC-SVM, \citep{Scholkopf:2001:ESH:1119748.1119749}) 
and support vector data description (SVDD, \cite{Tax:2004:SVD:960091.960109}) and rely on hand-crafted features. 
Recently, there is interest in deep learning methods for AD~\citep{DBLP:journals/corr/abs-1901-03407}. \citet{Erfani:2016:HLA:2952005.2952200,ae16,cao16hybrid,chen2017outlier} follow a hybrid approach where deep unsupervised learning is used to learn features, which are then used within a shallow AD method. Specifically for image domains, \citet{golan2018deep} learn features using a self-supervised paradigm -- by applying geometric transformations to the image and learning to classify which transformation was applied. Recently, self-supervised and contrastive approaches for image-based AD~\citep{tack2020csi, sehwag2021ssd} have exhibited impressive results by using augmentations or features from pre-trained models. \citet{lee2018simple} learn a distance metric in feature space, for networks pre-trained on image classification. All of these methods display outstanding AD performance, but are limited to image domains, while our approach does not require particular data or pre-trained features. Extending our approach to use unlabelled data, image augmentations, or pre-trained features is an interesting direction for future work. Several studies explored using deep generative models such as GANs and VAEs for AD~\citep{an2015variational,suh2016echo,schlegl2017unsupervised,wang2017safer}, and more recently, deep energy-based models \citep{DBLP:journals/corr/ZhaiCLZ16,Song2018LearningNR} have reported outstanding results.
Our work extends the VAE approach to the SSAD setting, showing that even a 1\% fraction of labelled anomalies can improve over the state-of-the-art AD scores of deep energy based models, demonstrating the importance of the SSAD setting.

Most studies on SSAD also require hand designed features. \citet{munoz2010semisupervised} proposed $S^2$OC-SVM, a modification of OC-SVM that introduces labeled and unlabeled samples, \citet{gornitz2013toward} proposed an approach based on SVDD, and \citet{Blanchard:2010:SND:1756006.1953028} base their approach on statistical testing. Also related is the active learning approach to AD~\citep{das16,das17,grill16}, which sequentially improves an anomaly detector by querying an expert for labels. Deep SSAD has been studied recently in specific contexts such as videos~\citep{kiran2018overview}, network intrusion detection~\citep{min2018ids}, or specific neural network architectures~\citep{ergen2017unsupervised}. 

The most relevant prior work to our study is the recently proposed deep semi-supervised anomaly detection (Deep SAD) approach of \citet{ruff2019}. Deep SAD is a general method based on deep SVDD~\citep{pmlr-v80-ruff18a}, which learns a neural-network mapping of the input that minimizes the volume of data around a predetermined point. While it has been shown to work on a variety of domains, Deep SAD must place restrictions on the network architecture, such as no bias terms (in all layers) and no bounded activation functions, to prevent degeneration of the minimization problem to a trivial solution. The methods we propose here do not place any restriction on the network architecture, and can be combined with any VAE model. Further, we show that the main strength of Deep SAD stems from its auto-encoder pre-training, making it effectively similar to a VAE-based approach. In this respect, our work provides a principled theoretical development of this approach.
In addition, we show improved performance in almost all domains compared to Deep SAD's state-of-the-art results.

\section{Background}
\label{bg}
% In this section, we provide background for our work. 
We consider deep unsupervised learning under the variational inference setting~\citep{kingma2013auto}. Given some data $x$, one aims to fit the parameters $\theta$ of a latent variable model $p_\theta(x) = \mathbb{E}_{p(z)} \left[p_\theta (x|z)\right]$, where the prior $p(z)$ is known. For general models, the maximum-likelihood objective $\max_\theta \log p_\theta(x)$ is intractable due to the marginalization over $z$, and can be approximated using variational inference methods.
\paragraph{Evidence lower bound (ELBO)}

The evidence lower bound (ELBO) states that for some approximate posterior distribution $q(z|x)$: 
% \begin{equation*}
$$
\log p_\theta(x) \geq \mathbb{E}_{q(z|x)}\ \left[\log p_\theta(x|z)\right] - \KL(q(z|x) \Vert p(z)), 
$$
% \end{equation*}
where the Kullback-Leibler divergence is
$\KL(q(z|x) \Vert p(z)) = \E_{q(z)} \left[ \log \frac{q(z|x)}{p(z)} \right]$. In the variational autoencoder (VAE,~\cite{kingma2013auto}), the approximate posterior is represented as $q_\phi(z|x) = \mathcal{N}(\mu_\phi(x), \Sigma_\phi(x))$ for some neural network with parameters $\phi$, the prior is $p(z) = \mathcal{N}(\mu_0,\Sigma_0)$, and the ELBO can be maximized using the \textit{reparameterization trick}. Since the resulting model resembles an autoencoder, the approximate posterior $q_\phi(z|x)$ is also known as the \emph{encoder}, while $p_\theta(x|z)$ is termed the \emph{decoder}. 

\paragraph{\texorpdfstring{$\chi$}{TEXT} upper bound (CUBO)}
Recently, \citet{dieng2017variational} derived variational upper bounds to the data log-likelihood. The $\chi^n$-divergence is given by 
$D_{\chi^n}(p \Vert q) = \E_{q(z;\theta)}\left[\left(\frac{p(z | X)}{q(z;\theta)} \right)^n - 1 \right].$
For $n>1$, \citet{dieng2017variational} propose the following $\chi$ upper bound (CUBO):
\small
% \begin{equation*}
$$
    \log p_\theta(x) \!\leq\! \frac{1}{n}\!\log \!\mathbf{E}_{q(z|x)}\! \!\left[\! \left( \!\frac{p_\theta(x|z)p(z)}{q(z|x)}\! \right)^n \right]\!\! \doteq \! CUBO_n(q(z|x)\!).
$$
% \end{equation*}
\normalsize
For a family of approximate posteriors $q_\phi(z|x)$, one can minimize the CUBO using Monte Carlo (MC) estimation. However, MC gives a lower bound to CUBO and its gradients are biased. As an alternative, \citet{dieng2017variational} proposed the following optimization objective: $ \mathcal{L} = \exp\{n \cdot CUBO_n(q_\phi(z|x))\} $. By monotonicity of $\exp$, this objective reaches the same optima as $CUBO_n(q_\phi(z|x))$. This produces an unbiased estimate, and the number of samples only affects the variance of the gradients. 

In the following, we denote by $ELBO_{Q}(X)$ and $CUBO_{Q}(X)$ the ELBO and CUBO for data $X$ where the approximate posterior is $Q(z|x)$.

\section{Deep Variational Semi-Supervised Anomaly Detection}
In the anomaly detection problem, the goal is to detect whether a sample $x$ was generated from some normal distribution $p_{normal}(x)$,\footnote{Throughout, subscript $\cdot_{normal}$ refers to the distribution of normal data, and not to the Gaussian distribution.} or not -- making it an anomaly. In semi-supervised anomaly detection (SSAD), we are given $N_{normal}$ samples from $p_{normal}(X)$, which we denote $X_{normal}$.\footnote{For clarity, our exposition assumes that $X_{normal}$ is clean, and does not contain any anomalous samples. We do verify that our methods can handle the important case of data polluted with anomalies in our experiments.} In addition, we are given $N_{outlier}$ examples of anomalous data, denoted by $X_{outlier}$, and we assume that $N_{outlier} \ll N_{normal}$. In particular, $X_{outlier}$ does not cover the range of possible anomalies, and thus classification methods (neither supervised nor semi-supervised) are not applicable.

Our approach for SSAD is to approximate $p_{normal}(x)$ using a deep latent variable model $p_{\theta}(x)$, and to decide whether a sample is anomalous or not based on thresholding its predicted likelihood. In the following, we propose two variational methods for learning $p_{\theta}(x)$. The first method, which we term \textit{max-min likelihood} VAE (MML-VAE), maximizes the likelihood of normal samples while minimizing the likelihood of outliers. The second method, which we term \textit{dual-prior} VAE (DP-VAE), assumes different priors for normal and outlier samples.
 
\subsection{Max-Min Likelihood VAE}
In this approach, we seek to find model parameters based on the following objective:
\begin{equation}\label{eq:max_min_objective}
% \begin{split}
    \max_{\theta} \quad \log p_{\theta}(X_{normal})- \gamma \log p_{\theta}(X_{outlier}),
% \end{split}
\end{equation}
where $\gamma \geq 0$ is a weighting term. 
Note that, in the absence of outlier data or when $\gamma = 0$, \Eqref{eq:max_min_objective} is just maximum likelihood estimation. For $\gamma > 0$, however, we take into account the knowledge that outliers should not be assigned high probability. 

We model the data distribution using a latent variable model $p_\theta(x) = \mathbb{E}_{p(z)} \left[p_\theta (x|z)\right]$, where the prior $p(z)$ is known. To optimize the objective in \Eqref{eq:max_min_objective} effectively, we propose the following variational lower bound:
\begin{equation}\label{eq:max_min_LB}
\begin{split}
   & \log p_{\theta}(X_{normal}) - \gamma\log p_{\theta}(X_{outlier}) \\
   & \geq  ELBO_{Q_1}(X_{normal}) - \gamma CUBO_{Q_2}(X_{outlier}),
\end{split}
\end{equation}
where $Q_1(z|x), Q_2(z|x)$ are the variational auxiliary distributions.
In principle, the objective in \Eqref{eq:max_min_LB} can be optimized using the methods of \citet{kingma2013auto} and \citet{dieng2017variational}, which would effectively require training two encoders: $Q_1$ and $Q_2$, and one decoder $p_\theta(x|z)$,\footnote{Based on the ELBO and CUBO definitions, the decoder $p_\theta(x|z)$ is the same for both terms. 
} separately on the two datasets. However, it is well-known that training deep generative models requires abundant data. Thus, there is little hope for learning an informative $Q_2$ using the small dataset $X_{outlier}$. To account for this, our main idea is to use the same variational distribution $Q(z|x)$ for both loss terms, which effectively relaxes the lower bound as follows: 
\begin{equation}\label{eq:max_min_LB_relaxed}
\begin{split}
& \log p_{\theta}(X_{normal}) - \gamma\log p_{\theta}(X_{outlier}) \\ & \geq ELBO_{Q}(X_{normal}) - \gamma CUBO_{Q}(X_{outlier}).
\end{split}
\end{equation}
In other words, we use the \textbf{same encoder} for both normal and anomalous data. \newtext{One can look at this design choice from a bias-variance perspective: sharing the encoder may loosen the lower bound (similar to adding bias), but in practice will work better when we have only a few negative samples (reducing variance).}
Finally, the loss function of the MML-VAE is: $ \mathcal{L} =  \gamma CUBO_{Q}(X_{outlier}) - ELBO_{Q}(X_{normal})$.

To gain intuition about the effect of the CUBO term in the loss function, it is instructive to assume that $p_\theta(x|z)$ is fixed. For $CUBO_2(q(z|x))$, the objective can be written as: $\mathbb{E}_q\left[\exp \left(-2 \log\left(\frac{q(z|x)}{p(x|z)p(z)} \right) \right) \right] $.
Minimizing this objective leads to pushing $q(z|x)$ away from $p(z)$. In this case, maximizing the lower bound only affects the variational distribution $q(z|x)$. 
% Note that the ELBO term seeks to minimize the KL distance between $q(z|X_{normal})$ and $p(z|X_{normal})$, which `pushes' $q(z|X_{normal})$ toward high-likelihood regions of $p(z)$, while the CUBO term is proportional to $\frac{1}{n}\log \E_{q(z|x)} \left[ \left( \frac{p(z|X_{outlier})}{q(z|X_{outlier})} \right)^n \right]$, and thus seeks to maximize the $\chi_n$ distance between $q(z|X_{outlier})$ and $p(z|X_{outlier})$. 
% \newtext{Now, assume that for some outlier, $p(z|X_{outlier})$ falls within a high-likelihood region of $p(z)$. In this case, the CUBO term will `push' $q(z|X_{outlier})$ away from $p(z)$.  
Thus, intuitively, the CUBO term seeks to \textbf{separate} the latent distribution for normal samples, which will concentrate on high-likelihood regions of $p(z)$, from the latents of outliers, which will concentrate elsewhere.  To further clarify, we plot the CUBO loss for a 1-dimensional Gaussian $q(z|x)\sim \mathcal{N}(\mu, \sigma)$, where the prior is $\mathcal{N}(0,1)$, and we fix $\mathcal{L}_R$. Observe that the CUBO loss drives $q(z|x)$ away from the prior to areas with $\mu > 0$ and small variance.

% \begin{figure*}[t]
% % \begin{SCfigure}
% % \begin{figure}
% \centering
% \includegraphics[width=5.5cm]{cubo_plot.png}
% \caption{CUBO loss for 1D Gaussian.}
% \label{wrap-fig:1}
% % \end{figure}
% % \end{SCfigure}
% \end{figure*}

For the common choice of Gaussian prior and approximate posterior distributions~\cite{kingma2013auto},  the CUBO component of the loss function is given by: 
% \begin{multline*} \mathcal{L}_{CUBO} =  \exp\{-2\cdot \mathcal{L}_R +  \log |\Sigma_q| +\mu_q^T\Sigma_q^{-1}\mu_q + \\ \log \E_{q}\big[\exp\{  -  z^Tz + z^T\Sigma_q^{-1}z -2z^T\Sigma_q^{-1}\mu_q  \} \big] \} .\end{multline*}
% \begin{equation*} 
$\mathcal{L}_{CUBO} =  \exp\{\log |\Sigma_q| +\mu_q^T\Sigma_q^{-1}\mu_q + \log \E_{q}\big[\exp\{ -2\cdot \mathcal{L}_R -  z^Tz + z^T\Sigma_q^{-1}z -2z^T\Sigma_q^{-1}\mu_q  \} \big] \},
$
% \end{equation*}
where $\mathcal{L}_R=-\log p_\theta(x|z)$ is the reconstruction error.
The expectation can be approximated with MC estimation. In our experiments, we found that computing $\exp({\log \E\left[\exp({\cdot)} \right]})$ is numerically more stable, as we can employ the log-sum-exp trick.

\begin{wrapfigure}{t}{5.5cm}
% \vspace{-1.5em}
\label{wrap-fig:1}
\includegraphics[width=5.5cm]{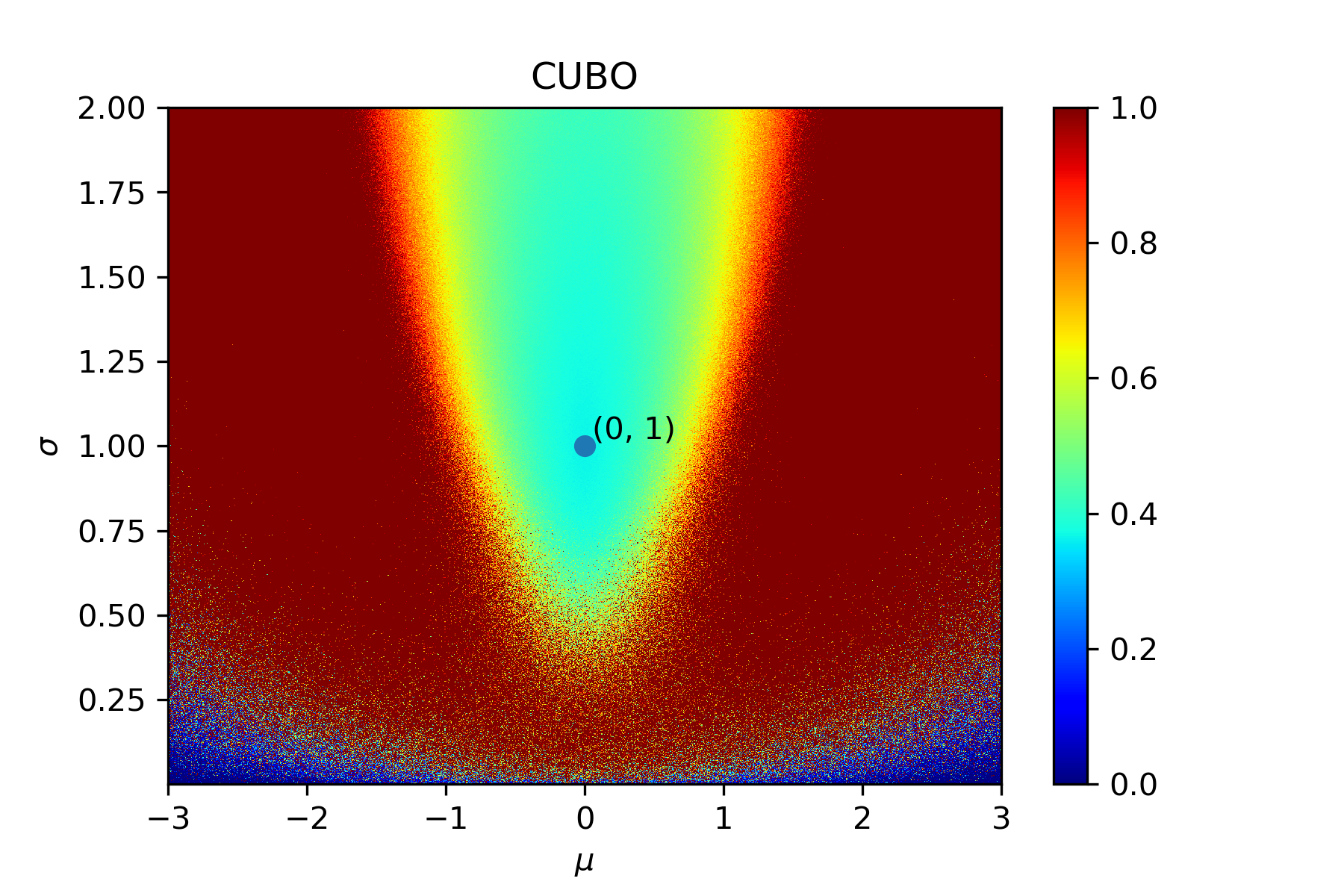}
\vspace{-1.0em}
\begin{small}
\caption{CUBO loss for 1D Gaussian.}
\end{small}
\vspace{-1.0em}
\end{wrapfigure}

In practice, we found that updating only the encoder according to the CUBO loss results in better performance and stabler training when the CUBO is estimated with a small number of samples, corresponding with the bias-variance intuition above. The reason for this is that CUBO seeks to maximize the reconstruction error, which does not produce informative updates for the decoder. Since this decoder is shared with the ELBO term, the CUBO update can decrease performance for normal samples as well. 
Complete algorithm details and derivation of the CUBO objective are provided
in Appendix \ref{cubo_deriv}.

\subsection{Dual Prior VAE}

The second method we propose is a simple yet effective modification of the VAE for SSAD, which we term dual-prior VAE (DP-VAE). Here, we assume that both normal and outlier data are generated from a single latent variable model $p_\theta(x) = \mathbb{E}_{p(z,y)} \left[p_\theta (x|z,y)\right]$, where $y$ is an additional Bernoulli variable that specifies whether the sample is normal or not. 
Additionally, we assume the following conditional prior for $z$:
% \footnote{This method can easily be extended to non-Gaussian priors, so long as the two priors are different.}
% \begin{equation*}
$$
    p(z|y) = \left\{\begin{array}{lr}
        \mathcal{N}(\mu_{normal}, \Sigma_{normal}), & y = 1\\
        \mathcal{N}(\mu_{outlier}, \Sigma_{outlier}), & y = 0
        \end{array}\right. .
$$
% \end{equation*}
Since our data is labeled, $y$ is known for every $x$ and does not require inference. Deriving the ELBO for normal and outlier samples in this case gives:
$ ELBO_{normal}(x) = -\mathcal{L}_R(x) - \KL[Q(z|x) \Vert p(z|y=1)],  $
$ ELBO_{outlier}(x) = -\mathcal{L}_R(x) - \KL[Q(z|x) \Vert p(z|y=0)].  $ 
Here, similarly to the MML-VAE above, we assume \textit{the same encoder} $Q(z|X)$ for normal and outlier samples, and the loss function is 
\small
$$\mathcal{L} = -\left[ ELBO_{normal}(X_{normal}) + ELBO_{outlier}(X_{outlier}) \right].$$
\normalsize
In our implementation, we set $\Sigma_{normal}=\Sigma_{outlier}=I$, $\mu_{normal}=0$, and $\mu_{outlier}= \alpha \vec{1}$ where $ \alpha \neq 0$. In this case, this loss function has the intuitive interpretation of training an encoder that separates between latent variables for normal and outlier data.  

As $\mathcal{L}$ is simply a combination of VAE loss functions, optimization is straightforward.
% \footnote{We also experimented with a hybrid model, trained by combining the loss functions for MML-VAE and DP-VAE, and can be seen as a DP-VAE with the additional loss of minimizing the likelihood of outlier samples in $ELBO_{normal}(x)$. This model obtained similar performance, as we report in Appendix \ref{tab:odds_results_appendix}.} 
Similarly to the MML-VAE method, we freeze the decoder when minimizing $ELBO_{outlier}$, as we don't expect for an informative reconstruction of the outlier distribution from the limited data $X_{outlier}$.

We emphasize that while this model assumes a prior distribution for the outliers (e.g., following the example in the introduction, for MNIST digits of class `1'), it can still be used in the SSAD setting to generalize to unseen outliers (digits from `1' to `9'). The reason, as we elaborate on in the next section, is that we will only use the normal data prior (for digit `0') for anomaly detection. Thus, similarly to the MML-VAE model, the DP-VAE works only to \textbf{separate} the latents of observed outliers from the latents of the normal data, therefore improving the detection of normal samples.

\subsection{VAE Architectures and Ensembles}
Both MML-VAE and DP-VAE simply add loss terms to the objective function of the conventional VAE. Thus, they can be applied to any type of VAE without any restriction on the architecture. In addition, they can be used with ensembles of VAEs, a popular technique to robustify VAE training and increase its performance in classification and generations tasks \citep{mocanu2018one, pmlr-v97-tan19b, shi2020dispersed, chen2017outlier,DBLP:journals/corr/abs-1802-09089}.
% \newtext{Preliminary results without the usage of ensembles, which are reported in the supplementary, have shown performance which is on par with the current SOTA, Deep SAD, but without the restrictions on the architectures.} 
We train $K$ different VAEs (either MML-VAE or DP-VAE) with random initial weights, and set the score to be the average over the ELBOs in the ensemble. We found that ensembles significantly improve the SSAD performance of our methods, 
% when supplied with outliers
as demonstrated in our experiments. In section \ref{sec:ablative} we further report an ablative analysis to evaluate the effect of using an ensemble.
% , and for a fair comparison with the state-of-the-art, we have also evaluated an ensemble of Deep SAD models, as we detail in Appendix \ref{img_full_res}.

\subsection{Anomaly Detection with MML-VAE and DP-VAE}
Finally, we describe our SSAD method based on MML-VAE and DP-VAE. After training an MML-VAE, the $ELBO$ provides an approximation of the normal data log-likelihood. Thus, given a novel sample $x$, we can score its (log) likelihood of belonging to normal data as:
% \begin{equation*}
$$
    score(x) = ELBO(x).
$$
% \end{equation*}
After training a DP-VAE, $ELBO_{normal}$ provides a similar approximation, and in this case we set 
$$score(x) = ELBO_{normal}(x).$$ 
Finally, for classifying the sample, we threshold its score.

\subsection{Comparison with SVDD and Deep SAD}\label{ss:comparison_deep_sad}
We discuss connections between our approaches and Deep SAD~\citep{ruff2019}, the recent state-of-the-art that builds on deep SVDD. 
In SVDD~\citep{pmlr-v80-ruff18a}, a deep network is trained to minimize the average distance in latent space between the features of the normal data samples and a predetermined point $c$, pushing the latent representations of normal data to concentrate in a hyper-sphere centered on $c$. The distance from $c$ is then used to distinguish normal samples (small distance) from outliers (large distance). While in principle $c$ could be arbitrary, in practice, the network is pre-trained using an auto-encoder loss, and $c$ is chosen by averaging the features of normal data. As we show in Appendix \ref{apndx_deepsad10}, choosing $c$ using the auto-encoder pre-training is crucial for SVDD performance (and thus, also for Deep SAD).
% -- when using these methods without fully pre-training the auto-encoder, performance drops drastically.
For SSAD, \citet{ruff2019} modify the SVDD objective to further push outliers away from $c$. While \citet{ruff2019} motivate this training using information theoretic terms, many steps in the explanation are not rigorously justified, such as the minimization of mutual information using the regularized auto-encoder, and using the inverse distance from $c$ when mapping features for outlier data. 

Interestingly, our methods, which are developed from a principled variational inference formulation, have a similar interpretation of learning features that minimize a (variational) auto-encoder loss, and `pushing away' the features of outliers from the features of normal samples. However, our auto-encoder loss is well-established as $p_\theta(x|z)$ in the ELBO. Further, our loss for outliers is developed rigorously, as a sequence of well-defined approximations to the data likelihood, in both methods.

% in both DP-VAE and MML-VAE.  

Furthermore, by building on well-established first principles, our method is not prone to the phenomena termed `hypersphere collapse', where the network can reduce the latent distance from $c$ to zero for all inputs. In order to avoid it, \citet{pmlr-v80-ruff18a,ruff2019} place strong restrictions on the model architecture (no bias terms, unbounded activations), or require modifications to the data~\citep{chong2020simple}. For the semi-supervised setting, \citet{ruff2019} notes that if there are sufficiently many labeled anomalies, hypersphere collapse is not a problem for Deep SAD. However, this is an empirical observation, and there is no result showing how many labeled anomalies are required to prevent hypersphere collapse. In practice, the restrictions of Deep SVDD were applied for obtaining the results in~\citet{ruff2019}.

We conclude with a note on ensembles. As we show in Section \ref{sec:experiments}, ensembles reduce the variance of the stochastic VAE training, improve our method’s performance, and lead to state-of-the-art results. For a fair comparison with Deep SAD, however, we have also evaluated ensembles of Deep SAD models, as reported in Appendix \ref{img_full_res}. Interestingly, our investigation shows that ensembles have little effect on Deep SAD. This can be explained as follows. In Deep-SAD, confidence is measured according to distance to an arbitrary point $c$ in the vector space. Thus, the scores from different networks in the ensemble are not necessarily calibrated (as they each have different $c$ points). In our VAE approach, however, the score of each network is derived to approximate the likelihood of the sample -- a calibrated quantity, naturally giving rise to the benefit of the ensemble.

\section{Experiments}\label{sec:experiments}

Our experiments aim to (1) establish our methods as SOTA for deep SSAD, following the extensive evaluation protocol of \citet{ruff2019}; and (2) demonstrate the generality of our approach for training VAEs with negative samples on various domains. The tasks and datasets are summarized in Table \ref{tab:exp_sum}, and note that due to lack of space, some results are reported in the appendix. In particular, our experiments 
on detecting novel sentiments in text data using a recurrent neural network (RNN) based VAE are in Appendix \ref{apndx_nlp}. 

\begin{table*}[ht]
\vspace{-1.0em}
\begin{center}
\begin{scriptsize}
\begin{tabular}{llcccccccc}
        \toprule
         \textbf{Domain} & \textbf{Task} & \textbf{Datasets}      & \textbf{Architecture}    \\
        \midrule
         				Images   & Anomaly Detection & MNIST, Fashion-MNIST, CIFAR10, CatsVsDogs      & CNN       \\
        	 				 Classic AD Benchmark \cite{Rayana:2016}   &     Anomaly Detection  &  arrhythmia, cardio, satellite, satimage-2, shuttle,  thyroid &  MLP  \\
        					 Robotics   &     Motion Planning  &      Path Planning \cite{ichter_mp_18}       & MLP \\
        	 				 NLP (Appendix \ref{apndx_nlp})  &     Sentiment  Analysis    & IMDB         &  RNN (LSTM)   \\
        \bottomrule
        \end{tabular}
\end{scriptsize}
\end{center}
\vspace{-0.5em}
\caption{Summary of the experimented tasks and data}
\label{tab:exp_sum}
\vspace{-1.0em}
\end{table*}

% \TD{Table \ref{tab:exp_sum} summarizes the tasks and data types we evaluate our methods on.}
% In our experiments, we follow the evaluation methods and datasets proposed in the extensive work of \cite{ruff2019}. 
\paragraph{SSAD Evaluation:} The SSAD evaluation protocol of \citet{ruff2019} includes strong baselines of shallow, deep, and hybrid algorithms for AD and SSAD. 
% Hybrid algorithms are defined as shallow SSAD methods that are trained on features extracted from a deep autoencoder trained on the raw data. 
A brief overview of the methods is provided in Appendix \ref{comp_methods}, and we refer the reader to \citet{ruff2019} for a more detailed description. Performance is evaluated by the area under the curve of the receiver operating characteristic curve (AUROC), a commonly used criterion for AD. There are two types of datasets: (1) high-dimensional datasets which were modified to be semi-supervised and (2) classic anomaly detection benchmark datasets. The first type includes MNIST, Fashion-MNIST and CIFAR-10, and the second includes datasets from various fields such as astronomy and medicine. For strengthening the baselines, \citet{ruff2019} grant the shallow and hybrid methods an unfair advantage of selecting their hyper-parameters to maximize AUROC on a subset (10\%) of the test set. Here we also follow this approach. 

\paragraph{Hyperparameters and Architectures:} When comparing with the state-of-the-art Deep SAD method, we used the same network architecture but include bias terms in all layers. For MNIST, we found that this architecture did not work well, and instead used a standard convolutional neural network (see Appendix \ref{s:implementation_details} for full details). 
Following the explanation in Section \ref{ss:comparison_deep_sad}, we use an ensemble of $K=5$ VAE models in all the experiments. We also report results without ensembling in Table \ref{tab:vs_ensemble} in the supplementary.
% \AT{Add some sentence on how you chose hyperparameters (maybe - Hyperparameters were chosen using a validation set that consisted of 20\% of the data?)} \TD{Moved the following part so here instead of the "image datasets" section}. 
Unlike \citet{ruff2019} which did not report on using a validation set to tune the hyper-parameters, we take a validation set (20\%) out of the training data to tune hyper-parameters for the image datasets.
% We use PyTorch \cite{paszke2017automatic} and a machine with an Nvidia RTX 2080 GPU.

\subsection{Image Datasets}
\paragraph{MNIST, Fashion-MNIST and CIFAR-10} datasets all have ten classes. Similarly to \citet{ruff2019}, we derive ten AD setups on each dataset. In each setup, one of the ten classes is set to be the normal class, and the remaining nine classes represent anomalies. During training, we set $X_{normal}$ as the normal class samples, and $X_{outlier}$ as a small fraction of the data from \textbf{only one} of the anomaly classes. At test time, we evaluate on anomalies from \textbf{all} classes. 

The experiment we perform, similarly to \citet{ruff2019}, is evaluating the model's detection ability as a function of the ratio of anomalies presented to it during training.
We set the ratio of labeled training data to be $\gamma_l = N_{outlier} / (N_{normal}+N_{outlier})$, and we evaluate different values of $\gamma_l$ in each scenario. In total, there are 90 experiments per each value of $\gamma_l$. Note that for $\gamma_l=0$, no labeled anomalies are presented, and we revert to standard \textit{unsupervised} AD, which in our approach amounts to training a standard VAE ensemble. For pre-processing, pixels are scaled to $[0,1]$.

Our results are presented in Table \ref{tab:img_results}. The complete table with all of the competing methods can be found in Appendix \ref{img_full_res}. Note that even a small fraction of labeled outliers ($\gamma_l=0.01$) leads to significant improvement compared to the standard unsupervised VAE ($\gamma_l=0$) and compared to the best-performing unsupervised methods on MNIST and CIFAR-10. Also, our methods consistently outperform DeepSAD and the other SSAD baselines in most domains, and the large standard deviation is a result of the high variability over the 90 different experiment settings. We validated that our improvement over Deep-SAD is statistically significant using a paired t-test ($p<0.05$).

% Our results are presented in Table \ref{tab:img_results}. The complete table with all of the competing methods can be found in Appendix \ref{img_full_res}. Note that even a small fraction of labeled outliers ($\gamma_l=0.01$) leads to significant improvement compared to the standard unsupervised VAE ($\gamma_l=0$) and compared to the best-performing unsupervised methods on MNIST and CIFAR-10. Also, our methods \TD{consistently} outperform \TD{DeepSAD and the} other SSAD baselines in most domains, \TD{and the large standard deviation is a result of the high variability over the 90 different experiment settings. We validated that our improvement over Deep-SAD is statistically significant using a paired t-test ($p<0.05$)}

Following \citet{ruff2019}, we also experiment with a setting where the training set is polluted with unknown outliers. The results, which are in Appendix \ref{img_full_res}, further show the robustness and effectiveness of our methods. 

\begin{table*}[ht]
\vspace{-0.5em}

\begin{center}
\begin{scriptsize}
\begin{tabular}{llcccccccccccccc}
        \toprule
                            &       & \textbf{OC-SVM}   & \textbf{OC-SVM}     & \textbf{Inclusive}  & \textbf{SSAD} & \textbf{SSAD}     &   \textbf{Deep}   & \textbf{Supervised} & \textbf{MML} & \textbf{DP} \\
        \textbf{Data}	    &$\gamma_l$ & \textbf{Raw}     & \textbf{Hybrid}   &  \textbf{NRF} & \textbf{Raw}  & \textbf{Hybrid}    &   \textbf{SAD}   & \textbf{Classifier} & \textbf{VAE} & \textbf{VAE}  \\
        \midrule
        MNIST 				& .00   & 96.0$\pm$2.9      & \textbf{96.3$\pm$2.5} &  95.2$\pm$3.0  & 96.0$\pm$2.9  & 96.3$\pm$2.5                  & 92.8$\pm$4.9 &  & 94.2$\pm$3.0 &  94.2$\pm$3.0  \\
        	 				& .01   &                   &  &  & 96.6$\pm$2.4  & 96.8$\pm$2.3      & 96.4$\pm$2.7 & 92.8$\pm$5.5 & \textbf{97.3$\pm$2.1} & 97.0$\pm$2.3  \\
        					& .05   &                   &                  & & 93.3$\pm$3.6  & 97.4$\pm$2.0      & 96.7$\pm$2.4 & 94.5$\pm$4.6  & \textbf{97.8$\pm$1.6} & 97.5$\pm$2.0   \\
        	 				& .10   &                   &                  &  & 90.7$\pm$4.4  & 97.6$\pm$1.7      & 96.9$\pm$2.3 & 95.0$\pm$4.7     & \textbf{97.8$\pm$1.6} & 97.6$\pm$2.1  \\
        	 				& .20   &                   &                &   & 87.2$\pm$5.6  & 97.8$\pm$1.5      & 96.9$\pm$2.4 & 95.6$\pm$4.4  & \textbf{97.9$\pm$1.6} & \textbf{97.9$\pm$1.8} \\
        \midrule
        F-MNIST             & .00   & \textbf{92.8$\pm$4.7}      & 91.2$\pm$4.7 &  & \textbf{92.8$\pm$4.7}  & 91.2$\pm$4.7  & 89.2$\pm$6.2 &      & 90.8$\pm$4.6 & 90.8$\pm$4.6          \\
        	 				& .01   &                   &                   & & \textbf{92.1$\pm$5.0}  & 89.4$\pm$6.0     & 90.0$\pm$6.4 & 74.4$\pm$13.6    & 91.2$\pm$6.6 & 90.9$\pm$6.7       \\
        					& .05   &                   &                  & & 88.3$\pm$6.2  & 90.5$\pm$5.9     & 90.5$\pm$6.5 & 76.8$\pm$13.2   & 91.6$\pm$6.3 & \textbf{92.2$\pm$4.6}        \\
        	 				& .10   &                   &                 &  & 85.5$\pm$7.1  & 91.0$\pm$5.6     & 91.3$\pm$6.0 & 79.0$\pm$12.3    & \textbf{91.7$\pm$6.4} & \textbf{91.7$\pm$6.0}        \\
        	 				& .20   &                   &                 &  &  82.0$\pm$8.0  & 89.7$\pm$6.6     & 91.0$\pm$5.5 & 81.4$\pm$12.0    & 91.9$\pm$6.0 & \textbf{92.1$\pm$5.7}       \\
        \midrule
        CIFAR-10	 		& .00   & 62.0$\pm$10.6     & 63.8$\pm$9.0  &  \textbf{70.0$\pm$4.9}& 62.0$\pm$10.6 & 63.8$\pm$9.0           & 60.9$\pm$9.4 & & 52.7$\pm$10.7 & 52.7$\pm$10.7  \\
        	 				& .01   &                   &                  & &  73.0$\pm$8.0  & 70.5$\pm$8.3       & 72.6$\pm$7.4 & 55.6$\pm$5.0      & 73.7$\pm$7.3 & \textbf{74.5$\pm$8.4}  \\
        					& .05   &                   &                 &  & 71.5$\pm$8.1  & 73.3$\pm$8.4      & 77.9$\pm$7.2 & 63.5$\pm$8.0    & \textbf{79.3$\pm$7.2} & 79.1$\pm$8.0       \\
        	 				& .10   &                   &                 &  & 70.1$\pm$8.1  & 74.0$\pm$8.1      & 79.8$\pm$7.1 & 67.7$\pm$9.6      & 80.8$\pm$7.7 & \textbf{81.1$\pm$8.1}      \\
        	 				& .20   &                   &                 &  & 67.4$\pm$8.8  & 74.5$\pm$8.0     & 81.9$\pm$7.0 & 80.5$\pm$5.9     & 82.6$\pm$7.2 & \textbf{82.8$\pm$7.3}  \\
        \bottomrule
        \end{tabular}
\end{scriptsize}
\end{center}
\vspace{-0.5em}
\caption{Results for image datasets.
We report the average and standard deviation of AUROC over 90 experiments, for various ratios of labeled anomalies in the data $\gamma_l$.}
\label{tab:img_results}
\end{table*}

\vspace{-1.0em}
\paragraph{CatsVsDogs Dataset}

In addition to the test domains of \citet{ruff2019}, we also evaluate on the CatsVsDogs dataset, which is notoriously difficult for anomaly detection~\citep{golan2018deep}. This dataset contains 25,000 images of cats and dogs in various positions, 12,500 in each class. Following \citet{golan2018deep}, We split this dataset into a training set containing 10,000 images, and a test set of 2,500 images in each class. We also rescale each image to size 64x64.  We follow a similar experimental procedure as described above, and average results over the two classes. 

We chose a VAE architecture similar to the autoencoder architecture in \citet{golan2018deep}, and for the  \textit{Deep SAD} baseline, we modified the architecture to not use bias terms and bounded activations. We report our results in Table \ref{tab:catsdogs_results}, along with the numerical scores for baselines taken from \citet{golan2018deep}.
Note that without labeled anomalies, our method is not informative, predicting roughly at chance level, and this aligns with the baseline results reported by \citet{golan2018deep}. However, \emph{even just $1\%$ labeled outliers is enough to significantly improve predictions and produce informative results}, demonstrating the potential of the SSAD. In this domain, the geometric transformations of \citet{golan2018deep} allow for significantly better performance even without labeled outliers. While this approach is domain specific, incorporating similar ideas into probabilistic AD methods is an interesting direction for future research.

\begin{table*}[ht]
%\vspace{-1.0em}
\begin{center}
\begin{small}
\begin{tabular}{llcccccccccccccc}
        \toprule
                            &       & \textbf{OC-SVM}   & \textbf{OC-SVM}      &      &     &    &   & \textbf{Deep} &\textbf{MML} & \textbf{DP} \\
        \textbf{Data}	    &$\gamma_l$ & \textbf{Raw}      & \textbf{Hybrid}    & \textbf{DAGMM}    &   \textbf{DSEBM}   & \textbf{ADGAN} & \textbf{DADGT}  & \textbf{SAD} & \textbf{VAE} & \textbf{VAE}  \\
        \midrule
        CatsVsDogs 				& .00   & 51.7      & 52.5    & 47.7                & 51.6 &  49.4 &  \textbf{88.8} & 49.9 & 50.7 & 50.7 \\
        	 				& .01   &                   &     &  &   &  & & 54.1 & 59.4 & \textbf{64.0}  \\
        					& .05   &                   &                &   &  & & & 60.1     & 68.4 & \textbf{70.3}    \\
        	 				& .10   &                   &                 &   &   &      &  & 64.4& 71.5 & \textbf{75.3}  \\
        	 				& .20   &                   &                   & &  &       &  & 67.2  & 73.3 & \textbf{78.2}  \\
        \bottomrule
        \end{tabular}
\end{small}
\end{center}
\vspace{-0.5em}
\caption{Results for ratio of labeled anomalies $\gamma_l$ in the training set, on the CatsVsDogs dataset.}
\vspace{-1.0em}
\label{tab:catsdogs_results}
\end{table*}

\subsection{Classic Anomaly Detection Datasets}
AD is a well-studied field of research, with many publicly available AD benchmarks and well-established baselines~\citep{Rayana:2016}. These datasets are lower-dimensional than the image datasets above, and by evaluating our method on them, we aim to demonstrate the flexibility of our method for general types of data. We follow \citet{ruff2019}, and consider random train-test split of 60:40 with stratified sampling to ensure correct representation of normal and anomaly classes. We evaluate over ten random seeds with 1\% of anomalies, i.e., $\gamma_l=0.01$. There are no specific anomaly classes, thus we treat all anomalies as one class. For pre-processing, we preform standardization of the features to have zero mean and unit variance. As most of these datasets have a very small amount of labeled anomalous data, it is inefficient to take a validation set from the training set. To tune hyper-parameters, we measure AUROC performance on the training data, and we follow \citet{ruff2019} and set a 150 training epochs limit. The complete hyper-parameters table is found in Appendix \ref{hyper_param}.

% \subsection{Results on the Classic Anomaly Detection datasets}
In Table \ref{tab:odds_results} we present the results of the best-performing algorithms along with our methods. The complete table with all the competing methods can be found in Appendix \ref{odds_full_res}. Our methods outperform the baselines on most datasets, demonstrating the flexibility of our approach.  
\begin{table*}[ht]
%\vspace{-1.0em}
\begin{center}
\begin{small}
\begin{tabular}{lcccccccc}
\toprule
                    & \textbf{OC-SVM}         & \textbf{SSAD}          & \textbf{Supervised}       & \textbf{Deep} &  & \textbf{MML} & \textbf{DP}  \\
\textbf{Dataset}    & \textbf{Raw} 	    & \textbf{Raw}               & \textbf{Classifier}       & \textbf{SAD} & \textbf{VAE} & \textbf{VAE} & \textbf{VAE} \\
\midrule
arrhythmia          & 84.5$\pm$3.9           & \textbf{86.7$\pm$4.0}          & 39.2$\pm$9.5              & 75.9$\pm$8.7 & 85.6 $\pm$ 2.4 & 85.7 $\pm$ 2.3 & \textbf{86.7 $\pm$ 1.7}  \\
cardio              & 98.5$\pm$0.3           & 98.8$\pm$0.3        & 83.2$\pm$9.6              & 95.0$\pm$1.6 & 95.5 $\pm$ 0.8 & \textbf{99.2 $\pm$ 0.4} & 99.1 $\pm$ 0.4  & \\
satellite           & 95.1$\pm$0.2          & \textbf{96.2$\pm$0.3}      & 87.2$\pm$2.1              & 91.5$\pm$1.1  & 77.7 $\pm$ 1.0  & 92.0 $\pm$ 1.5  & 89.2 $\pm$ 1.6  \\
satimage-2          & 99.4$\pm$0.8          & \textbf{99.9$\pm$0.1}         & \textbf{99.9$\pm$0.1}     & \textbf{99.9$\pm$0.1}  & 99.6 $\pm$ 0.6  & 99.7 $\pm$ 0.3  & \textbf{99.9 $\pm$ 0.1}  \\
shuttle             & 99.4$\pm$0.9          & 99.6$\pm$0.5          & 95.1$\pm$8.0              & 98.4$\pm$0.9  & 98.1 $\pm$ 0.6  & 99.8 $\pm$ 0.1  &  \textbf{99.9 $\pm$ 0.04}  \\
thyroid             & 98.3$\pm$0.9           & 97.9$\pm$1.9             & 97.8$\pm$2.6              & 98.6$\pm$0.9  & 86.9 $\pm$ 3.7  & \textbf{99.9 $\pm$ 0.04} & \textbf{99.9 $\pm$ 0.04} \\
\bottomrule
\end{tabular}
\end{small}
\end{center}
% \vspace{-0.5em}
\caption{Results on classic AD benchmark datasets in the setting with a ratio of labeled anomalies of $\gamma_l = 0.01$ in the training set. We report the avg. AUROC with st.dev.~computed over 10 seeds.}
\label{tab:odds_results}
\end{table*}

\vspace{-1.5em}
\subsection{Ablative Analysis}
\vspace{-1.0em}
\label{sec:ablative}
We perform an ablative analysis of DP-VAE method on the cardio, satellite and CIFAR-10 datasets (with 1\% outliers). We evaluate our method with the following properties: (1) Frozen and unfrozen decoder, (2) separate and same encoder for normal data and outliers and (3) the effect of using ensembles of VAEs instead of one.
Table \ref{tab:ablation_an} summarizes the analysis; the complete table can be found in Appendix \ref{apndx_ab_an}. It can be seen that using the same encoder is necessary as we expected, since training a neural network requires sufficient amount of data. Moreover, when dealing with a small pool of outliers, the effect on the decoder is minimal. Hence, freezing the decoder contributes little to the improvement. 
% Ensembles improve our method’s result by approximately 2-4\% on average, which makes sense as VAE training is a stochastic process. For a fair comparison with the SOTA, we have also evaluated an ensemble of Deep SAD models. These results, which are detailed in Appendix \ref{img_full_res}, show that ensembles have little effect on Deep SAD. This can be explained as follows. In Deep-SAD, confidence is measured according to distance to an arbitrary point $c$ in the vector space. Thus, the scores from different networks in the ensemble are not necessarily calibrated (they have different $c$ points). In our VAE approach, however, the score of each network is derived to approximate the likelihood of the sample -- a calibrated quantity, giving rise to the benefit of the ensemble.

\begin{figure*}[!ht]
    % \vspace{-5.0em}
    \centering
    \includegraphics[width=\textwidth]{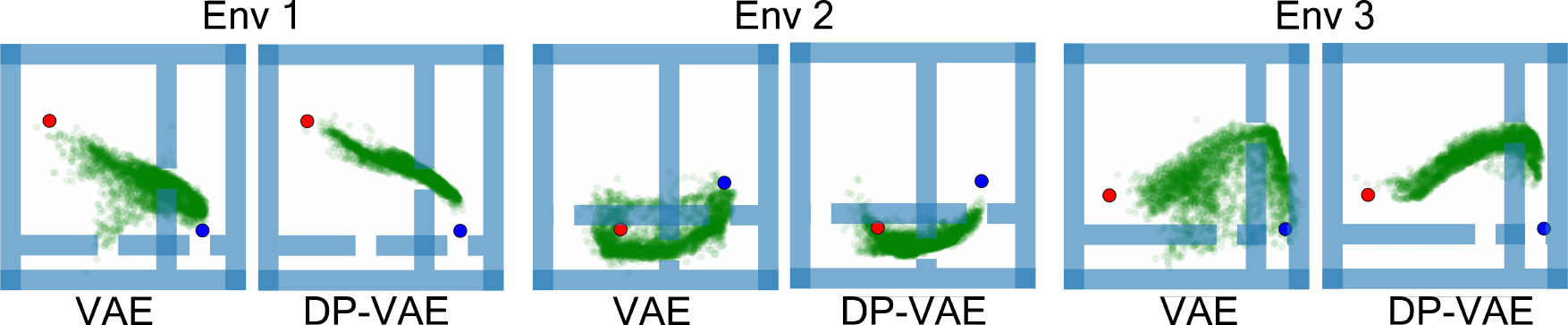}
    \caption{Motion planning with VAE. Following the work of \citet{ichter_mp_18}, a conditional VAE is trained to generate robot configurations to be used within a sampling based motion planning algorithm. The VAE generates 6D robot configurations (robot position and velocity; here projected onto the 2D plane) given an image of obstacles, an initial state (blue) and a goal state (red). Using our DP-VAE method, we added to the VAE training negative samples on obstacle boundaries. We compare samples from the standard VAE (left) and DP-VAE (right) on three unseen obstacle maps. Note that our method results in much more informative samples for planning obstacle avoidance.}
    \label{fig:mp}
    \vspace{-1em}
\end{figure*}

\begin{table*}[h!]
% \vspace{-8.0em}
\begin{center}
\begin{small}
\begin{tabular}{llcccccccc}
        \toprule
                                               &                       &                   &   \textbf{Cardio} & \textbf{Satellite} & \textbf{CIFAR-10} \\
         \textbf{Encoder} & \textbf{Decoder}      & \textbf{Ensemble}   & \textbf{AUROC} & \textbf{AUROC} & \textbf{AUROC}  \\
        \midrule
         				Separate   & Unfreeze      & 5     & 96.6$\pm$0.6 & 65.3$\pm$1.1 & 49.8$\pm$10.7   \\
        	 				 Separate   &     Freeze              &  5 & 96.6$\pm$0.6  & 66.9$\pm$1.2 & 50.0$\pm$10.9  \\
        					 Same   &     Unfreeze              &  5 & 98.8$\pm$0.06 & 88.9$\pm$1.7  & 73.9$\pm$8.9 \\
        	 				 Same   &     Freeze              &  5 & \textbf{99.1$\pm$0.4} & \textbf{91.7$\pm$1.5} & \textbf{74.5$\pm$8.4}   \\
        	 				 Same   &     Freeze              &  1 & 97.8$\pm$1.1 & 87.6$\pm$1.7 & 72.3$\pm$9.0  \\
        \bottomrule
        \end{tabular}
\end{small}
\end{center}
\vspace{-0.5em}
\caption{Ablative analysis (DP-VAE). AUROC is reported over average of 10 seeds for the \textit{satellite} and \textit{cardio}. For \textit{CIFAR-10}, results are reported for 1\% of anomalies, averaged over 90 experiments.}
\label{tab:ablation_an}
\vspace{-1.0em}
\end{table*}

\vspace{-0.5em}
\subsection{Sample-Based Motion Planning Application}
\vspace{-1.0em}
While our focus is on SSAD, our methods can be used to enhance any VAE generative model when negative samples are available. We demonstrate this idea in a motion planning domain~\citep{latombe2012robot}. Sampling-based planning algorithms search for a path for a robot between obstacles on a graph of points sampled from the feasible configurations of the system. 
% mostly rely on uniform sampling strategies. 
In order to accelerate the planning process, \citet{ichter_mp_18} suggest to learn non-uniform sampling strategies that concentrate in regions where an optimal solution might lie, given a training set of planning problems to learn from. \citet{ichter_mp_18} proposed a conditional VAE that samples configurations conditioned on an image of the obstacles in the domain, and is trained using samples of feasible paths on a set of training domains.

Here, we propose to enhance the quality of the samples by including a few outliers (5\% of the normal data) during training, which we choose to be points on the obstacle boundaries -- as such points clearly do not belong to the motion plan. 
We used the publicly available code base of \citet{ichter_mp_18}, and modified only the CVAE training using our DP-VAE method. Exemplary generated samples for different obstacle configurations (unseen during training) are shown in Figure \ref{fig:mp}. It can be seen that adding the outliers led to a VAE that is much more focused on the feasible parts of the space, and thus generates significantly less useless points in collision.

\vspace{-0.5em}
\section{Conclusion}
\vspace{-1.0em}
We proposed two VAE modifications that account for negative data examples, and used them for semi-supervised anomaly detection. We showed that these methods can be derived from natural probabilistic formulations of the problem, and that the resulting algorithms are general and effective -- competitive or better than the state-of-the-art on diverse datasets. We further demonstrated that even a small fraction of outlier data can significantly improve anomaly detection on various datasets, and that our methods can be combined with VAE applications such as in motion planning and NLP.

We see great potential in the probabilistic approach to AD using deep generative models: it has a principled probabilistic interpretation, it is agnostic to the particular type of data, and it can be implemented using expressive generative models. For specific data such as images, however, discriminative approaches that exploit domain specific methods such as geometric transformations are currently the best performers. Developing similar self-supervised methods for generative approaches is an exciting direction for future research. 

\section{Acknowledgements}
\vspace{-1.0em}
This work is partly funded by the Israel Science Foundation (ISF-759/19) and the Open Philanthropy Project Fund, an advised fund of Silicon Valley Community Foundation.

{\small
\bibliographystyle{plainnat}
\bibliography{refs}

\begin{thebibliography}{51}
\providecommand{\natexlab}[1]{#1}
\providecommand{\url}[1]{\texttt{#1}}
\expandafter\ifx\csname urlstyle\endcsname\relax
  \providecommand{\doi}[1]{doi: #1}\else
  \providecommand{\doi}{doi: \begingroup \urlstyle{rm}\Url}\fi

\bibitem[An and Cho(2015)]{an2015variational}
Jinwon An and Sungzoon Cho.
\newblock Variational autoencoder based anomaly detection using reconstruction
  probability.
\newblock \emph{Special Lecture on IE}, 2\penalty0 (1), 2015.

\bibitem[Andrews et~al.(2016)Andrews, Morton, and Griffin]{ae16}
Jerone Andrews, Edward Morton, and Lewis Griffin.
\newblock Detecting anomalous data using auto-encoders.
\newblock \emph{International Journal of Machine Learning and Computing},
  6:\penalty0 21, 01 2016.

\bibitem[Blanchard et~al.(2010)Blanchard, Lee, and
  Scott]{Blanchard:2010:SND:1756006.1953028}
Gilles Blanchard, Gyemin Lee, and Clayton Scott.
\newblock Semi-supervised novelty detection.
\newblock \emph{J. Mach. Learn. Res.}, 11:\penalty0 2973--3009, December 2010.

\bibitem[Bowman et~al.(2016)Bowman, Vilnis, Vinyals, Dai, Jozefowicz, and
  Bengio]{bowman-etal-2016-generating}
Samuel~R. Bowman, Luke Vilnis, Oriol Vinyals, Andrew Dai, Rafal Jozefowicz, and
  Samy Bengio.
\newblock Generating sentences from a continuous space.
\newblock In \emph{Proceedings of The 20th {SIGNLL} Conference on Computational
  Natural Language Learning}, pages 10--21, Berlin, Germany, August 2016.
  Association for Computational Linguistics.
\newblock \doi{10.18653/v1/K16-1002}.
\newblock URL \url{https://www.aclweb.org/anthology/K16-1002}.

\bibitem[Cao et~al.(2016)Cao, Nicolau, and Mcdermott]{cao16hybrid}
Van~Loi Cao, Miguel Nicolau, and James Mcdermott.
\newblock A hybrid autoencoder and density estimation model for anomaly
  detection.
\newblock In \emph{PPSN}, volume 9921, pages 717--726, 09 2016.

\bibitem[Chalapathy and Chawla(2019)]{DBLP:journals/corr/abs-1901-03407}
Raghavendra Chalapathy and Sanjay Chawla.
\newblock Deep learning for anomaly detection: {A} survey.
\newblock \emph{CoRR}, abs/1901.03407, 2019.

\bibitem[Chen et~al.(2017)Chen, Sathe, Aggarwal, and Turaga]{chen2017outlier}
Jinghui Chen, Saket Sathe, Charu Aggarwal, and Deepak Turaga.
\newblock Outlier detection with autoencoder ensembles.
\newblock In \emph{Proceedings of the 2017 SIAM International Conference on
  Data Mining}, pages 90--98. SIAM, 2017.

\bibitem[Chong et~al.(2020)Chong, Ruff, Kloft, and Binder]{chong2020simple}
Penny Chong, Lukas Ruff, Marius Kloft, and Alexander Binder.
\newblock Simple and effective prevention of mode collapse in deep one-class
  classification, 2020.

\bibitem[Daniel and Tamar(2021)]{daniel2021soft}
Tal Daniel and Aviv Tamar.
\newblock Soft-introvae: Analyzing and improving the introspective variational
  autoencoder.
\newblock In \emph{Proceedings of the IEEE/CVF Conference on Computer Vision
  and Pattern Recognition}, pages 4391--4400, 2021.

\bibitem[{Das} et~al.(2016){Das}, {Wong}, {Dietterich}, {Fern}, and
  {Emmott}]{das16}
S.~{Das}, W.~{Wong}, T.~{Dietterich}, A.~{Fern}, and A.~{Emmott}.
\newblock Incorporating expert feedback into active anomaly discovery.
\newblock In \emph{2016 IEEE 16th International Conference on Data Mining
  (ICDM)}, pages 853--858, 2016.

\bibitem[Das et~al.(2017)Das, Wong, Fern, Dietterich, and Siddiqui]{das17}
Shubhomoy Das, Weng{-}Keen Wong, Alan Fern, Thomas~G. Dietterich, and Md~Amran
  Siddiqui.
\newblock Incorporating feedback into tree-based anomaly detection.
\newblock \emph{CoRR}, abs/1708.09441, 2017.

\bibitem[Deecke et~al.(2018)Deecke, Vandermeulen, Ruff, Mandt, and
  Kloft]{deecke2018anomaly}
Lucas Deecke, Robert Vandermeulen, Lukas Ruff, Stephan Mandt, and Marius Kloft.
\newblock Anomaly detection with generative adversarial networks, 2018.

\bibitem[Dieng et~al.(2017)Dieng, Tran, Ranganath, Paisley, and
  Blei]{dieng2017variational}
Adji~Bousso Dieng, Dustin Tran, Rajesh Ranganath, John Paisley, and David Blei.
\newblock Variational inference via $\chi $ upper bound minimization.
\newblock In \emph{Advances in Neural Information Processing Systems}, pages
  2732--2741, 2017.

\bibitem[Erfani et~al.(2016)Erfani, Rajasegarar, Karunasekera, and
  Leckie]{Erfani:2016:HLA:2952005.2952200}
Sarah~M. Erfani, Sutharshan Rajasegarar, Shanika Karunasekera, and Christopher
  Leckie.
\newblock High-dimensional and large-scale anomaly detection using a linear
  one-class svm with deep learning.
\newblock \emph{Pattern Recogn.}, 58\penalty0 (C):\penalty0 121--134, October
  2016.

\bibitem[Ergen et~al.(2017)Ergen, Mirza, and Kozat]{ergen2017unsupervised}
Tolga Ergen, Ali~Hassan Mirza, and Suleyman~Serdar Kozat.
\newblock Unsupervised and semi-supervised anomaly detection with lstm neural
  networks.
\newblock \emph{arXiv preprint arXiv:1710.09207}, 2017.

\bibitem[Golan and El-Yaniv(2018)]{golan2018deep}
Izhak Golan and Ran El-Yaniv.
\newblock Deep anomaly detection using geometric transformations.
\newblock In \emph{Advances in Neural Information Processing Systems}, pages
  9758--9769, 2018.

\bibitem[Goodfellow et~al.(2014)Goodfellow, Pouget-Abadie, Mirza, Xu,
  Warde-Farley, Ozair, Courville, and Bengio]{goodfellow2014generative}
Ian Goodfellow, Jean Pouget-Abadie, Mehdi Mirza, Bing Xu, David Warde-Farley,
  Sherjil Ozair, Aaron Courville, and Yoshua Bengio.
\newblock Generative adversarial nets.
\newblock In \emph{Advances in neural information processing systems}, pages
  2672--2680, 2014.

\bibitem[G{\"o}rnitz et~al.(2013)G{\"o}rnitz, Kloft, Rieck, and
  Brefeld]{gornitz2013toward}
Nico G{\"o}rnitz, Marius Kloft, Konrad Rieck, and Ulf Brefeld.
\newblock Toward supervised anomaly detection.
\newblock \emph{Journal of Artificial Intelligence Research}, 46:\penalty0
  235--262, 2013.

\bibitem[Grill and Pevn\'{y}(2016)]{grill16}
Martin Grill and Tom\'{a}\v{s} Pevn\'{y}.
\newblock Learning combination of anomaly detectors for security domain.
\newblock \emph{Comput. Netw.}, 107\penalty0 (P1):\penalty0 55–63, October
  2016.
\newblock ISSN 1389-1286.
\newblock \doi{10.1016/j.comnet.2016.05.021}.

\bibitem[{Ichter} et~al.(2018){Ichter}, {Harrison}, and {Pavone}]{ichter_mp_18}
B.~{Ichter}, J.~{Harrison}, and M.~{Pavone}.
\newblock Learning sampling distributions for robot motion planning.
\newblock In \emph{2018 IEEE International Conference on Robotics and
  Automation (ICRA)}, pages 7087--7094, May 2018.

\bibitem[{Kaae S{\o}nderby} et~al.(2016){Kaae S{\o}nderby}, {Raiko},
  {Maal{\o}e}, {Kaae S{\o}nderby}, and {Winther}]{2016arXiv160202282K}
Casper {Kaae S{\o}nderby}, Tapani {Raiko}, Lars {Maal{\o}e}, S{\o}ren {Kaae
  S{\o}nderby}, and Ole {Winther}.
\newblock {Ladder Variational Autoencoders}.
\newblock \emph{arXiv e-prints}, art. arXiv:1602.02282, Feb 2016.

\bibitem[Kingma and Ba(2014)]{adam_14}
Diederik Kingma and Jimmy Ba.
\newblock Adam: A method for stochastic optimization.
\newblock \emph{International Conference on Learning Representations}, 12 2014.

\bibitem[Kingma and Welling(2013)]{kingma2013auto}
Diederik~P Kingma and Max Welling.
\newblock Auto-encoding variational bayes.
\newblock \emph{arXiv preprint arXiv:1312.6114}, 2013.

\bibitem[Kingma et~al.(2014)Kingma, Rezende, Mohamed, and
  Welling]{DBLP:journals/corr/KingmaRMW14}
Diederik~P. Kingma, Danilo~Jimenez Rezende, Shakir Mohamed, and Max Welling.
\newblock Semi-supervised learning with deep generative models.
\newblock \emph{CoRR}, abs/1406.5298, 2014.

\bibitem[Kiran et~al.(2018)Kiran, Thomas, and Parakkal]{kiran2018overview}
B~Kiran, Dilip Thomas, and Ranjith Parakkal.
\newblock An overview of deep learning based methods for unsupervised and
  semi-supervised anomaly detection in videos.
\newblock \emph{Journal of Imaging}, 4\penalty0 (2):\penalty0 36, 2018.

\bibitem[Latombe(2012)]{latombe2012robot}
Jean-Claude Latombe.
\newblock \emph{Robot motion planning}, volume 124.
\newblock Springer Science \& Business Media, 2012.

\bibitem[Lee et~al.(2018)Lee, Lee, Lee, and Shin]{lee2018simple}
Kimin Lee, Kibok Lee, Honglak Lee, and Jinwoo Shin.
\newblock A simple unified framework for detecting out-of-distribution samples
  and adversarial attacks.
\newblock In \emph{Advances in Neural Information Processing Systems}, pages
  7167--7177, 2018.

\bibitem[Liu et~al.(2008)Liu, Ting, and Zhou]{liu2008isolation}
Fei~Tony Liu, Kai~Ming Ting, and Zhi-Hua Zhou.
\newblock Isolation forest.
\newblock In \emph{2008 Eighth IEEE International Conference on Data Mining},
  pages 413--422. IEEE, 2008.

\bibitem[Min et~al.(2018)Min, Long, Liu, Cui, Cai, and Ma]{min2018ids}
Erxue Min, Jun Long, Qiang Liu, Jianjing Cui, Zhiping Cai, and Junbo Ma.
\newblock Su-ids: A semi-supervised and unsupervised framework for network
  intrusion detection.
\newblock In \emph{International Conference on Cloud Computing and Security},
  pages 322--334. Springer, 2018.

\bibitem[Mirsky et~al.(2018)Mirsky, Doitshman, Elovici, and
  Shabtai]{DBLP:journals/corr/abs-1802-09089}
Yisroel Mirsky, Tomer Doitshman, Yuval Elovici, and Asaf Shabtai.
\newblock Kitsune: An ensemble of autoencoders for online network intrusion
  detection.
\newblock \emph{CoRR}, abs/1802.09089, 2018.

\bibitem[Mocanu and Mocanu(2018)]{mocanu2018one}
Decebal~Constantin Mocanu and Elena Mocanu.
\newblock One-shot learning using mixture of variational autoencoders: a
  generalization learning approach.
\newblock \emph{arXiv preprint arXiv:1804.07645}, 2018.

\bibitem[Mu{\~n}oz-Mar{\'\i} et~al.(2010)Mu{\~n}oz-Mar{\'\i}, Bovolo,
  G{\'o}mez-Chova, Bruzzone, and Camp-Valls]{munoz2010semisupervised}
Jordi Mu{\~n}oz-Mar{\'\i}, Francesca Bovolo, Luis G{\'o}mez-Chova, Lorenzo
  Bruzzone, and Gustavo Camp-Valls.
\newblock Semisupervised one-class support vector machines for classification
  of remote sensing data.
\newblock \emph{IEEE transactions on geoscience and remote sensing},
  48\penalty0 (8):\penalty0 3188--3197, 2010.

\bibitem[Nalisnick et~al.(2019)Nalisnick, Matsukawa, Teh, Gorur, and
  Lakshminarayanan]{nalisnick2018do}
Eric Nalisnick, Akihiro Matsukawa, Yee~Whye Teh, Dilan Gorur, and Balaji
  Lakshminarayanan.
\newblock Do deep generative models know what they don't know?
\newblock In \emph{International Conference on Learning Representations}, 2019.

\bibitem[Pennington et~al.(2014)Pennington, Socher, and
  Manning]{Pennington14glove:global}
Jeffrey Pennington, Richard Socher, and Christopher~D. Manning.
\newblock Glove: Global vectors for word representation.
\newblock In \emph{In EMNLP}, 2014.

\bibitem[Pimentel et~al.(2014)Pimentel, Clifton, Clifton, and
  Tarassenko]{pimentel2014review}
Marco~AF Pimentel, David~A Clifton, Lei Clifton, and Lionel Tarassenko.
\newblock A review of novelty detection.
\newblock \emph{Signal Processing}, 99:\penalty0 215--249, 2014.

\bibitem[Ruff et~al.(2018)Ruff, Vandermeulen, G{\"o}rnitz, Deecke, Siddiqui,
  Binder, M{\"u}ller, and Kloft]{pmlr-v80-ruff18a}
Lukas Ruff, Robert~A. Vandermeulen, Nico G{\"o}rnitz, Lucas Deecke, Shoaib~A.
  Siddiqui, Alexander Binder, Emmanuel M{\"u}ller, and Marius Kloft.
\newblock Deep one-class classification.
\newblock In \emph{Proceedings of the 35th International Conference on Machine
  Learning}, volume~80, pages 4393--4402, 2018.

\bibitem[Ruff et~al.(2020)Ruff, Vandermeulen, G{\"o}rnitz, Binder, M{\"u}ller,
  M{\"u}ller, and Kloft]{ruff2019}
Lukas Ruff, Robert~A. Vandermeulen, Nico G{\"o}rnitz, Alexander Binder,
  Emmanuel M{\"u}ller, Klaus-Robert M{\"u}ller, and Marius Kloft.
\newblock Deep semi-supervised anomaly detection.
\newblock In \emph{ICLR}, 2020.

\bibitem[Schlegl et~al.(2017)Schlegl, Seeb{\"o}ck, Waldstein, Schmidt-Erfurth,
  and Langs]{schlegl2017unsupervised}
Thomas Schlegl, Philipp Seeb{\"o}ck, Sebastian~M Waldstein, Ursula
  Schmidt-Erfurth, and Georg Langs.
\newblock Unsupervised anomaly detection with generative adversarial networks
  to guide marker discovery.
\newblock In \emph{International Conference on Information Processing in
  Medical Imaging}, pages 146--157. Springer, 2017.

\bibitem[Sch\"{o}lkopf et~al.(2001)Sch\"{o}lkopf, Platt, Shawe-Taylor, Smola,
  and Williamson]{Scholkopf:2001:ESH:1119748.1119749}
Bernhard Sch\"{o}lkopf, John~C. Platt, John~C. Shawe-Taylor, Alex~J. Smola, and
  Robert~C. Williamson.
\newblock Estimating the support of a high-dimensional distribution.
\newblock \emph{Neural Comput.}, 13\penalty0 (7):\penalty0 1443--1471, July
  2001.

\bibitem[Sehwag et~al.(2021)Sehwag, Chiang, and Mittal]{sehwag2021ssd}
Vikash Sehwag, Mung Chiang, and Prateek Mittal.
\newblock Ssd: A unified framework for self-supervised outlier detection.
\newblock \emph{arXiv preprint arXiv:2103.12051}, 2021.

\bibitem[Shebuti(2016)]{Rayana:2016}
Rayana Shebuti.
\newblock Odds library, 2016.
\newblock URL \url{http://odds.cs.stonybrook.edu}.

\bibitem[Shi et~al.(2020)Shi, Zhou, Miao, and Li]{shi2020dispersed}
Wenxian Shi, Hao Zhou, Ning Miao, and Lei Li.
\newblock Dispersed exponential family mixture vaes for interpretable text
  generation, 2020.

\bibitem[Song and Ou(2018)]{Song2018LearningNR}
Yunfu Song and Zhijian Ou.
\newblock Learning neural random fields with inclusive auxiliary generators.
\newblock \emph{ArXiv}, abs/1806.00271, 2018.

\bibitem[Suh et~al.(2016)Suh, Chae, Kang, and Choi]{suh2016echo}
Suwon Suh, Daniel~H Chae, Hyon-Goo Kang, and Seungjin Choi.
\newblock Echo-state conditional variational autoencoder for anomaly detection.
\newblock In \emph{2016 International Joint Conference on Neural Networks
  (IJCNN)}, pages 1015--1022. IEEE, 2016.

\bibitem[Tack et~al.(2020)Tack, Mo, Jeong, and Shin]{tack2020csi}
Jihoon Tack, Sangwoo Mo, Jongheon Jeong, and Jinwoo Shin.
\newblock Csi: Novelty detection via contrastive learning on distributionally
  shifted instances.
\newblock \emph{arXiv preprint arXiv:2007.08176}, 2020.

\bibitem[Tan and Peharz(2019)]{pmlr-v97-tan19b}
Ping~Liang Tan and Robert Peharz.
\newblock Hierarchical decompositional mixtures of variational autoencoders,
  09--15 Jun 2019.

\bibitem[Tax and Duin(2004)]{Tax:2004:SVD:960091.960109}
David M.~J. Tax and Robert P.~W. Duin.
\newblock Support vector data description.
\newblock \emph{Mach. Learn.}, 54\penalty0 (1):\penalty0 45--66, January 2004.
\newblock ISSN 0885-6125.

\bibitem[Wang et~al.(2017)Wang, Wang, Tamar, Chen, and Abbeel]{wang2017safer}
William Wang, Angelina Wang, Aviv Tamar, Xi~Chen, and Pieter Abbeel.
\newblock Safer classification by synthesis.
\newblock \emph{CoRR}, abs/1711.08534, 2017.

\bibitem[Zhai et~al.(2016)Zhai, Cheng, Lu, and
  Zhang]{DBLP:journals/corr/ZhaiCLZ16}
Shuangfei Zhai, Yu~Cheng, Weining Lu, and Zhongfei Zhang.
\newblock Deep structured energy based models for anomaly detection.
\newblock \emph{CoRR}, abs/1605.07717, 2016.

\bibitem[Zisselman and Tamar(2020)]{Zisselman_2020_CVPR}
Ev~Zisselman and Aviv Tamar.
\newblock Deep residual flow for out of distribution detection.
\newblock In \emph{The IEEE Conference on Computer Vision and Pattern
  Recognition (CVPR)}, June 2020.

\bibitem[Zong et~al.(2018)Zong, Song, Min, Cheng, Lumezanu, Cho, and
  Chen]{zong2018deep}
Bo~Zong, Qi~Song, Martin~Renqiang Min, Wei Cheng, Cristian Lumezanu, Daeki Cho,
  and Haifeng Chen.
\newblock Deep autoencoding gaussian mixture model for unsupervised anomaly
  detection.
\newblock In \emph{International Conference on Learning Representations}, 2018.

\end{thebibliography}
}
\clearpage

%%%%%%%%%%%%%%%%%%%%%%%%%%%%%%%%%%%%%%%%%%%%%%%%%%%%%%%%%%%%

\appendix

% \section{Appendix}

\section{Complete Results}
\subsection{MNIST, Fashion-MNIST, CIFAR-10}
\label{img_full_res}
Table \ref{tab:img_full_res} includes the complete results. The results for an ensemble of Deep SAD models are in parenthesis (CIFAR-10). The ensemble method is implemented the same way as ours: we train $K=5$ separate models (i.e., each model has its own $c$), and the score (which is the distance from $c$) is the average of scores from all models in the ensemble. 

 \begin{table}[ht]
%\vspace{-1.0em}
\begin{center}
\begin{normalsize}
\begin{tabular}{llcccccccc}
        \toprule
                                          \textbf{Anomaly}        &      \textbf{Deep} & &\textbf{MML} && \textbf{DP} &  \\
         \textbf{Ratio} & \textbf{SAD} && \textbf{VAE} && \textbf{VAE} &       \\
         \midrule & $K=1$ & $K=5$ & $K=1$ & $K=5$ & $K=1$ & $K=5$ \\
        \midrule
         				0.0   &  60.9$\pm$9.4 & 60.7$\pm$9.4 & 56.4$\pm$6.5 & 52.7$\pm$10.7 & 56.4$\pm$6.5 & 52.7$\pm$10.7  \\
        	 				0.01   & 72.6$\pm$7.4 & 71.7$\pm$8.4 & 72.6 $\pm$8.7 & 73.7$\pm$7.3 & 72.5$\pm$8.2 & 74.5$\pm$8.4\\
        	 				0.05   & 77.9$\pm$7.2  & 77.7$\pm$7.8 & 77.1 $\pm$ 7.6 & 79.3$\pm$7.2 & 77.3$\pm$8.3& 79.1$\pm$8.0 \\
        	 				0.1   & 79.8$\pm$7.1  & 79.6$\pm$7.4 & 78.7 $\pm$ 7.7 & 80.8$\pm$7.7 & 79.1$\pm$ 8.4 & 81.1$\pm$8.1 \\
        	 				0.2   &  81.9$\pm$7.0 & 81.2$\pm$7.5 & 80.5$\pm$ 7.3 & 82.6$\pm$7.2 & 80.4$\pm$8.4 & 82.8$\pm$7.3\\
        \bottomrule
        \end{tabular}
\end{normalsize}
\end{center}
\caption{CIFAR-10 AUROC comparison between ensemble size $K$ of Deep SAD, MML-VAE and DP-VAE models.  We report the avg. AUROC with st.dev. computed over 90 experiments at various ratios $\gamma_l$.}
\label{tab:vs_ensemble}
\vspace{-0.5em}
\end{table}

Table \ref{tab:img_pol_results_appendix} includes the complete results for the setting with a ratio of $\gamma_p$ polluted samples. We follow \citet{ruff2019} and further investigate the robustness of our method in a scenario where the training set is polluted with unknown outliers. We fix the ratio of labeled training samples at $\gamma_l = 0.05$ and similarly to the previous experiment we only draw known outliers form one anomaly class. The results show that our methods outperform Deep SAD most of the time or show competitive performance.
\afterpage{%
    \clearpage%
    \begin{landscape}% Landscape page
        \centering % Center table
        \begin{table}[ht]
        \begin{tiny}
        %%%%%%%%%%%%%%%%%%%%%%%%%%%%%%%%%%%%%%%%
        \setlength{\tabcolsep}{0.7em}
        \def\arraystretch{1.5}
        \begin{tabular}{llcccccccccccccccc}
        \toprule
                            &       & \textbf{OC-SVM}   & \textbf{OC-SVM}   & \textbf{IF}   & \textbf{IF}       & \textbf{KDE}      & \textbf{KDE}      &                   & \textbf{Deep}  \\
        \textbf{Data}	    &$\gamma_l$ & \textbf{Raw}      & \textbf{Hybrid}   & \textbf{Raw}  & \textbf{Hybrid}   & \textbf{Raw}      & \textbf{Hybrid}   & \textbf{CAE}      & \textbf{SVDD}  \\
        \midrule
        MNIST 				& .00   & 96.0$\pm$2.9      & 96.3$\pm$2.5      & 85.4$\pm$8.7  & 90.5$\pm$5.3      & 95.0$\pm$3.3      & 87.8$\pm$5.6      & 92.9$\pm$5.7      & 92.8$\pm$4.9    \\
        	 				& .01   &                   &                   &               &                   &                   &                   &                   &                  \\
        					& .05   &                   &                   &               &                   &                   &                   &                   &                   \\
        	 				& .10   &                   &                   &               &                   &                   &                   &              &      \\
        	 				& .20   &                   &                   &               &                   &                   &                   &                   &                 \\
        \midrule
        F-MNIST             & .00   & 92.8$\pm$4.7      & 91.2$\pm$4.7      & 91.6$\pm$5.5  & 82.5$\pm$8.1      & 92.0$\pm$4.9      & 69.7$\pm$14.4     & 90.2$\pm$5.8      & 89.2$\pm$6.2       \\
        	 				& .01   &                   &                   &               &                   &                   &                   &                   &                \\
        					& .05   &                   &                   &               &                   &                   &                   &                   &                 \\
        	 				& .10   &                   &                   &               &                   &                   &                   &                   &                 \\
        	 				& .20   &                   &                   &               &                   &                   &                   &              &     \\
        \midrule
        CIFAR-10	 		& .00   & 62.0$\pm$10.6     & 63.8$\pm$9.0      & 60.0$\pm$10.0 & 59.9$\pm$6.7      & 59.9$\pm$11.7     & 56.1$\pm$10.2     & 56.2$\pm$13.2     & 60.9$\pm$9.4         \\
        	 				& .01   &                   &                   &               &                   &                   &                   &                   &                   \\
        					& .05   &                   &                   &               &                   &                   &                   &                   &                 \\
        	 				& .10   &                   &                   &               &                   &                   &                   &                   &                 \\
        	 				& .20   &                   &                   &               &                   &                   &                   &                   &                  \\
        \bottomrule
        \end{tabular}
        
        \begin{tabular}{llcccccccccccccccc}
        \toprule
                            &       &  \textbf{Inclusive}     & \textbf{SSAD} & \textbf{SSAD}     &                   &   \textbf{Deep}   & \textbf{Supervised} & \textbf{MML} & \textbf{DP}  \\
        \textbf{Data}	    &$\gamma_l$ &  \textbf{NRF}     & \textbf{Raw}  & \textbf{Hybrid}   & \textbf{SS-DGM}   &   \textbf{SAD}   & \textbf{Classifier} & \textbf{VAE} & \textbf{VAE} \\
        \midrule
        MNIST 				& .00   &  95.26$\pm$3.0     & 96.0$\pm$2.9  & 96.3$\pm$2.5      &                   & 92.8$\pm$4.9 &  & 94.2$\pm$3.0 &  94.2$\pm$3.0    \\
        	 				& .01  & & 96.6$\pm$2.4  & 96.8$\pm$2.3      & 89.9$\pm$9.2      & 96.4$\pm$2.7 & 92.8$\pm$5.5 & 97.3$\pm$2.1 & 97.0$\pm$2.3              \\
        					& .05   & & 93.3$\pm$3.6  & 97.4$\pm$2.0      & 92.2$\pm$5.6      & 96.7$\pm$2.4 & 94.5$\pm$4.6  & 97.8$\pm$1.6 & 97.5$\pm$2.0   \\
        	 				& .10   & & 90.7$\pm$4.4  & 97.6$\pm$1.7      & 91.6$\pm$5.5      & 96.9$\pm$2.3 & 95.0$\pm$4.7     & 97.8$\pm$1.6 & 97.6$\pm$2.1   \\
        	 				& .20   &  & 87.2$\pm$5.6  & 97.8$\pm$1.5      & 91.2$\pm$5.6      & 96.9$\pm$2.4 & 95.6$\pm$4.4  & 97.9$\pm$1.6 & 97.9$\pm$1.8   \\
        \midrule
        F-MNIST             & .00  & & 92.8$\pm$4.7  & 91.2$\pm$4.7      &                   & 89.2$\pm$6.2 &      & 90.8$\pm$4.6 & 90.8$\pm$4.6   \\
        	 				& .01   &   & 92.1$\pm$5.0  & 89.4$\pm$6.0      & 65.1$\pm$16.3     & 90.0$\pm$6.4 & 74.4$\pm$13.6    & 91.2$\pm$6.6 & 90.9$\pm$6.7  \\
        					& .05  &   & 88.3$\pm$6.2  & 90.5$\pm$5.9      & 71.4$\pm$12.7     & 90.5$\pm$6.5 & 76.8$\pm$13.2   & 91.6$\pm$6.3 & 92.2$\pm$4.6 \\
        	 				& .10  &  & 85.5$\pm$7.1  & 91.0$\pm$5.6      & 72.9$\pm$12.2     & 91.3$\pm$6.0 & 79.0$\pm$12.3    & 91.7$\pm$6.4 & 91.7$\pm$6.0   \\
        	 				& .20  &       & 82.0$\pm$8.0  & 89.7$\pm$6.6      & 74.7$\pm$13.5     & 91.0$\pm$5.5 & 81.4$\pm$12.0    & 91.9$\pm$6.0 & 92.1$\pm$5.7  \\
        \midrule
        CIFAR-10	 		& .00   & 70.0$\pm$4.9  & 62.0$\pm$10.6 & 63.8$\pm$9.0      &                   & 60.9$\pm$9.4 [60.7$\pm$9.4] & & 52.7$\pm$10.7 & 52.7$\pm$10.7         \\
        	 				& .01  & & 73.0$\pm$8.0  & 70.5$\pm$8.3      & 49.7$\pm$1.7      & 72.6$\pm$7.4 [71.7$\pm$8.4]& 55.6$\pm$5.0      & 73.7$\pm$7.3 & 74.5$\pm$8.4  \\
        					& .05   &  & 71.5$\pm$8.1  & 73.3$\pm$8.4      & 50.8$\pm$4.7      & 77.9$\pm$7.2 [77.7$\pm$7.8] & 63.5$\pm$8.0    & 79.3$\pm$7.2 & 79.1$\pm$8.0 \\
        	 				& .10  &  & 70.1$\pm$8.1  & 74.0$\pm$8.1      & 52.0$\pm$5.5      & 79.8$\pm$7.1 [79.6$\pm$7.4] & 67.7$\pm$9.6      & 80.8$\pm$7.7 & 81.1$\pm$8.1   \\
        	 				& .20  &  & 67.4$\pm$8.8  & 74.5$\pm$8.0      & 53.2$\pm$6.7      & 81.9$\pm$7.0 [81.2$\pm$7.5] & 80.5$\pm$5.9     & 82.6$\pm$7.2 & 82.8$\pm$7.3  \\
        \bottomrule
        \end{tabular}
        
        %%%%%%%%%%%%%%%%%%%%%%%%%%%%%%%%%%%%%%%%
        \end{tiny}
        \caption[Complete image datasets results]{Complete results of the experiment where we increase the ratio of labeled anomalies $\gamma_l$ in the training set. We report the avg. AUROC with st.dev. computed over 90 experiments at various ratios $\gamma_l$. In parenthesis, the results of an ensemble of Deep SAD models ($K=5$).}
        \label{tab:img_full_res}
        \end{table}
        
    \end{landscape}
    \clearpage% Flush page
}

\subsection{Classic Anomaly Detection}
\label{odds_full_res}
Table \ref{tab:odds_results_appendix} includes the complete results. We also experimented with a hybrid model, trained by combining the loss functions for MML-VAE and DP-VAE, and can be seen as a DP-VAE with the additional loss of minimizing the likelihood of outlier samples in $ELBO_{normal}(x)$. This model obtained similar performance.
\afterpage{%
    \clearpage%
    \begin{landscape}% Landscape page
        \centering % Center table
        \begin{scriptsize}
        %%%%%%%%%%%%%%%%%%%%%%%%%%%%%%%%%%%%%%%%
        \begin{tabular}{lccccccccccccc}
        \toprule
                            & \textbf{OC-SVM}   & \textbf{OC-SVM}       &                       & \textbf{Deep}     & \textbf{SSAD}     & \textbf{SSAD}         &                   & \textbf{Deep}     & \textbf{Supervised} &  & \textbf{MML} & \textbf{DP} & \textbf{MML-DP} \\
        \textbf{Data}	    & \textbf{Raw}      & \textbf{Hybrid}       & \textbf{CAE}          & \textbf{SVDD}     & \textbf{Raw}      & \textbf{Hybrid}       & \textbf{SS-DGM}   & \textbf{SAD}      & \textbf{Classifier} & \textbf{VAE} & \textbf{VAE} & \textbf{VAE} & \textbf{VAE} \\
        \midrule
        arrhythmia          & 84.5$\pm$3.9      & 76.7$\pm$6.2          & 74.0$\pm$7.5          & 74.6$\pm$9.0      & 86.7$\pm$4.0      & 78.3$\pm$5.1          & 50.3$\pm$9.8      & 75.9$\pm$8.7      & 39.2$\pm$9.5 & 85.6 $\pm$ 2.4 & 85.7 $\pm$ 2.3 & 86.7 $\pm$ 1.7 & 87.3 $\pm$ 1.7 \\
        cardio              & 98.5$\pm$0.3      & 82.8$\pm$9.3          & 94.3$\pm$2.0          & 84.8$\pm$3.6      & 98.8$\pm$0.3      & 86.3$\pm$5.8          & 66.2$\pm$14.3     & 95.0$\pm$1.6      & 83.2$\pm$9.6 & 95.5 $\pm$ 0.8 & 99.3 $\pm$ 0.3 & 99.1 $\pm$ 0.4  & 99.2 $\pm$ 0.4 \\
        satellite           & 95.1$\pm$0.2      & 68.6$\pm$4.8          & 80.0$\pm$1.7          & 79.8$\pm$4.1      & 96.2$\pm$0.3      & 86.9$\pm$2.8          & 57.4$\pm$6.4      & 91.5$\pm$1.1      & 87.2$\pm$2.1 & 77.7 $\pm$ 1.0  & 92.0 $\pm$ 1.5  & 89.2 $\pm$ 1.6  & 91.7 $\pm$ 1.5 \\
        satimage-2          & 99.4$\pm$0.8      & 96.7$\pm$2.1          & 99.9$\pm$0.0          & 98.3$\pm$1.4      & 99.9$\pm$0.1      & 96.8$\pm$2.1          & 99.2$\pm$0.6      & 99.9$\pm$0.1      & 99.9$\pm$0.1 & 99.6 $\pm$ 0.6  & 99.7 $\pm$ 0.3  & 99.9 $\pm$ 0.1  & 99.8 $\pm$ 0.1 \\
        shuttle             & 99.4$\pm$0.9      & 94.1$\pm$9.5          & 98.2$\pm$1.2          & 86.3$\pm$7.5      & 99.6$\pm$0.5      & 97.7$\pm$1.0          & 97.9$\pm$0.3      & 98.4$\pm$0.9      & 95.1$\pm$8.0 & 98.1 $\pm$ 0.6  & 99.8 $\pm$ 0.1  &  99.9 $\pm$ 0.04  & 99.9$ \pm$ 0.09\\
        thyroid             & 98.3$\pm$0.9      & 91.2$\pm$4.0          & 75.2$\pm$10.2         & 72.0$\pm$9.7      & 97.9$\pm$1.9      & 95.3$\pm$3.1          & 72.7$\pm$12.0     & 98.6$\pm$0.9      & 97.8$\pm$2.6 & 86.9 $\pm$ 3.7  & 99.9 $\pm$ 0.04 & 99.9 $\pm$ 0.04 & 99.9 $\pm$ 0.03 \\
        \bottomrule
        \end{tabular}
        %%%%%%%%%%%%%%%%%%%%%%%%%%%%%%%%%%%%%%%%
        \end{scriptsize}
        \captionsetup{type=table}\caption{Complete results on classic AD benchmark datasets in the setting with  a ratio of labeled anomalies of $\gamma_l = 0.01$ in the training set. We report the avg. AUROC with st.dev. computed over 10 seeds.}
        \label{tab:odds_results_appendix}
    \end{landscape}
    \clearpage% Flush page
}

%%%%%%%%%%%%%%%%%%%%%%%%%%%%%%%%%%%%%%%%%%%%%%%%%%%%%%%%%%%%%%%%%%%%%%%%%%%%%%%%

\afterpage{%
    \clearpage%
    \begin{landscape}% Landscape page
        \centering % Center table
        \begin{table}[ht]
        \begin{scriptsize}
        \begin{tabular}{llccccccccccccccc}
        \toprule
                            &       & \textbf{OC-SVM}   & \textbf{OC-SVM}   & \textbf{IF}           & \textbf{IF}           & \textbf{KDE}          & \textbf{KDE}      &                   & \textbf{Deep}     \\
        \textbf{Data}	    & $\gamma_p$& \textbf{Raw}      & \textbf{Hybrid}   & \textbf{Raw}          & \textbf{Hybrid}       & \textbf{Raw}          & \textbf{Hybrid}   & \textbf{CAE}      & \textbf{SVDD}     \\
        \midrule
        MNIST 				& .00 	& 96.0$\pm$2.9      & 96.3$\pm$2.5      & 85.4$\pm$8.7          & 90.5$\pm$5.3          & 95.0$\pm$3.3          & 87.8$\pm$5.6      & 92.9$\pm$5.7      & 92.8$\pm$4.9       \\
        	 				& .01 	& 94.3$\pm$3.9      & 95.6$\pm$2.5      & 85.2$\pm$8.8          & 90.6$\pm$5.0          & 91.2$\pm$4.9          & 87.9$\pm$5.3      & 91.3$\pm$6.1      & 92.1$\pm$5.1      \\
        					& .05 	& 91.4$\pm$5.2      & 93.8$\pm$3.9      & 83.9$\pm$9.2          & 89.7$\pm$6.0          & 85.5$\pm$7.1          & 87.3$\pm$7.0      & 87.2$\pm$7.1      & 89.4$\pm$5.8      \\
        	 				& .10 	& 88.8$\pm$6.0      & 91.4$\pm$5.1      & 82.3$\pm$9.5          & 88.2$\pm$6.5          & 82.1$\pm$8.5          & 85.9$\pm$6.6      & 83.7$\pm$8.4      & 86.5$\pm$6.8      \\
        	 				& .20 	& 84.1$\pm$7.6      & 85.9$\pm$7.6      & 78.7$\pm$10.5         & 85.3$\pm$7.9          & 77.4$\pm$10.9         & 82.6$\pm$8.6      & 78.6$\pm$10.3     & 81.5$\pm$8.4     \\
        \midrule
        F-MNIST             & .00	& 92.8$\pm$4.7      & 91.2$\pm$4.7      & 91.6$\pm$5.5          & 82.5$\pm$8.1          & 92.0$\pm$4.9          & 69.7$\pm$14.4     & 90.2$\pm$5.8      & 89.2$\pm$6.2     \\
        	 				& .01 	& 91.7$\pm$5.0      & 91.5$\pm$4.6      & 91.5$\pm$5.5          & 84.9$\pm$7.2          & 89.4$\pm$6.3          & 73.9$\pm$12.4     & 87.1$\pm$7.3      & 86.3$\pm$6.3      \\
        					& .05 	& 90.7$\pm$5.5      & 90.7$\pm$4.9      & \textbf{90.9$\pm$5.9}          & 85.5$\pm$7.2          & 85.2$\pm$9.1          & 75.4$\pm$12.9     & 81.6$\pm$9.6      & 80.6$\pm$7.1      \\
        	 				& .10 	& 89.5$\pm$6.1      & 89.3$\pm$6.2      & \textbf{90.2$\pm$6.3}          & 85.5$\pm$7.7          & 81.8$\pm$11.2         & 77.8$\pm$12.0     & 77.4$\pm$11.1     & 76.2$\pm$7.3      \\
        	 				& .20 	& 86.3$\pm$7.7      & 88.1$\pm$6.9      & 88.4$\pm$7.6          & 86.3$\pm$7.4          & 77.4$\pm$13.6         & 82.1$\pm$9.8      & 72.5$\pm$12.6     & 69.3$\pm$6.3      \\
        \midrule
        CIFAR-10	 		& .00 	& 62.0$\pm$10.6     & 63.8$\pm$9.0      & 60.0$\pm$10.0         & 59.9$\pm$6.7          & 59.9$\pm$11.7         & 56.1$\pm$10.2     & 56.2$\pm$13.2     & 60.9$\pm$9.4     \\
        	 				& .01 	& 61.9$\pm$10.6     & 63.8$\pm$9.3      & 59.9$\pm$10.1         & 59.9$\pm$6.7          & 59.2$\pm$12.3         & 56.3$\pm$10.4     & 56.2$\pm$13.1     & 60.5$\pm$9.4     \\
        					& .05 	& 61.4$\pm$10.7     & 62.6$\pm$9.2      & 59.6$\pm$10.1         & 59.6$\pm$6.4          & 58.1$\pm$12.9         & 55.6$\pm$10.5     & 55.7$\pm$13.3     & 59.6$\pm$9.8      \\
        	 				& .10 	& 60.8$\pm$10.7     & 62.9$\pm$8.2      & 58.8$\pm$10.1         & 59.1$\pm$6.6          & 57.3$\pm$13.5         & 54.9$\pm$11.1     & 55.4$\pm$13.3     & 58.6$\pm$10.0     \\
        	 				& .20 	& 60.3$\pm$10.3     & 61.9$\pm$8.1      & 57.9$\pm$10.1         & 58.3$\pm$6.2          & 56.2$\pm$13.9         & 54.2$\pm$11.1     & 54.6$\pm$13.3     & 57.0$\pm$10.6     \\
        \bottomrule
        \end{tabular}
        
        %%%%%%%%%%%%%%%%%%%%%%%%%%%%%%%%%%%%%%%%
        \begin{tabular}{llccccccccccccccc}
        \toprule
                            &       &  \textbf{SSAD}     & \textbf{SSAD}     &                   & \textbf{Deep}     & \textbf{Supervised} & \textbf{MML} & \textbf{DP} \\
        \textbf{Data}	    & $\gamma_p$&  \textbf{Raw}      & \textbf{Hybrid}   & \textbf{SS-DGM}   & \textbf{SAD}      & \textbf{Classifier} & \textbf{VAE} & \textbf{VAE} \\
        \midrule
        MNIST 				& .00 	&  \textbf{97.9$\pm$1.8}      & 97.4$\pm$2.0      & 92.2$\pm$5.6      & 96.7$\pm$2.4      & 94.5$\pm$4.6 & 97.8$\pm$1.6 & 97.5$\pm$2.0 \\
        	 				& .01 	&  \textbf{96.6$\pm$2.4}      & 95.2$\pm$2.3      & 92.0$\pm$6.0      & 95.5$\pm$3.3      & 91.5$\pm$5.9 & 94.7$\pm$4.2 & 95.7$\pm$4.0\\
        					& .05 	& 93.4$\pm$3.4      & 89.5$\pm$3.9      & 91.0$\pm$6.9      & 93.5$\pm$4.1      & 86.7$\pm$7.4 & 91.8$\pm$5.3 & \textbf{93.9$\pm$4.8}\\
        	 				& .10 	&  90.7$\pm$4.4      & 86.0$\pm$4.6      & 89.7$\pm$7.5      & 91.2$\pm$4.9      & 83.6$\pm$8.2 & 89.3$\pm$6.0 & \textbf{92.3$\pm$5.4}\\
        	 				& .20 	& 87.4$\pm$8.6      & 86.6$\pm$6.6      & 79.7$\pm$9.4 & 85.4$\pm$7.7 & \textbf{88.8$\pm$6.3}\\
        \midrule
        F-MNIST             & .00	&  \textbf{94.0$\pm$4.4}      & 90.5$\pm$5.9      & 71.4$\pm$12.7     & 90.5$\pm$6.5      & 76.8$\pm$13.2 & 91.6$\pm$6.3 & 92.2$\pm$4.6\\
        	 				& .01 	& \textbf{92.2$\pm$4.9 }     & 87.8$\pm$6.1      & 71.2$\pm$14.3     & 87.2$\pm$7.1      & 67.3$\pm$8.1 & 88.1$\pm$9.8 & 85.7$\pm$10.6\\
        					& .05 	&  88.3$\pm$6.2      & 82.7$\pm$7.8      & 71.9$\pm$14.3     & 81.5$\pm$8.5      & 59.8$\pm$4.6 & 89.8$\pm$7.5 & 80.3$\pm$13.4\\
        	 				& .10 	&  85.6$\pm$7.0      & 79.8$\pm$9.0      & 72.5$\pm$15.5     & 78.2$\pm$9.1      & 56.7$\pm$4.1 & 88.9$\pm$11.2 & 75.9$\pm$15.7\\
        	 				& .20 	&  81.9$\pm$8.1      & 74.3$\pm$10.6     & 70.8$\pm$16.0     & 74.8$\pm$9.4      & 53.9$\pm$2.9 & \textbf{89.2$\pm$8.3} & 71.5$\pm$17.4\\
        \midrule
        CIFAR-10	 		& .00 	&  73.8$\pm$7.6      & 73.3$\pm$8.4      & 50.8$\pm$4.7      & 77.9$\pm$7.2      & 63.5$\pm$8.0 & \textbf{79.3$\pm$7.2} & 79.1$\pm$8.0\\
        	 				& .01 	&  73.0$\pm$8.0      & 72.8$\pm$8.1      & 51.1$\pm$4.7      & 76.5$\pm$7.2      & 62.9$\pm$7.3 & 76.9$\pm$8.2 & \textbf{78.5$\pm$8.7}\\
        					& .05 	& 71.5$\pm$8.2      & 71.0$\pm$8.4      & 50.1$\pm$2.9      & 74.0$\pm$6.9      & 62.2$\pm$8.2 & 75.3$\pm$8.3 & \textbf{76.8$\pm$8.9}\\
        	 				& .10 	&  69.8$\pm$8.4      & 69.3$\pm$8.5      & 50.5$\pm$3.6      & 71.8$\pm$7.0      & 60.6$\pm$8.3 & 73.5$\pm$8.9 & \textbf{75.2$\pm$9.2}\\
        	 				& .20 	&  67.8$\pm$8.6      & 67.9$\pm$8.1      & 50.1$\pm$1.7      & 68.5$\pm$7.1      & 58.5$\pm$6.7 & 71.0$\pm$9.6 & \textbf{72.2$\pm$10.3}\\
        \bottomrule
        \end{tabular}
        %%%%%%%%%%%%%%%%%%%%%%%%%%%%%%%%%%%%%%%%
        
        \end{scriptsize}
        \caption[Complete results of the pollution experiment]{Complete results of the setting where we pollute the unlabeled part of the training set with (unknown) anomalies. We report the avg. AUROC with st.~dev.~computed over 90 experiments at various ratios $\gamma_p$.}
        \label{tab:img_pol_results_appendix}
        \end{table}
    \end{landscape}
    \clearpage% Flush page
}

\section{Competing Methods}
\label{comp_methods}
We compare our methods to reported performance of both deep and shallow learning approaches, as detailed by \citet{ruff2019} and \citet{golan2018deep}. For completeness, we give a brief overview of the methods.

\paragraph{OC-SVM/SVDD} The one-class support vector machine (OC-SVM) is a kernel based method for novelty detection 
\citep{Scholkopf:2001:ESH:1119748.1119749}. It is typically employed with an RBF kernel, and learns
a collection of closed sets in the input space, containing most of the training samples. SVDD \citep{Tax:2004:SVD:960091.960109} is equivalent to to OC-SVM for the RBF kernel. OC-SVM are both granted an unfair advantage by selecting its hyper-parameters to maximize the AUROC on a subset (10\%) of the test set to establish a strong baseline.

\paragraph{Isolation Forest (IF)} Proposed by \citet{liu2008isolation}, IF is a tree-based method that explicitly isolates anomalies instead of constructing a profile of normal instances and then identify instances that do not conform to the normal profile as anomalies. As recommended in the original paper, the number of trees is set to $t=100$, and the sub-sampling size to $\psi = 256$.

\paragraph{Kernel Density Estimator (KDE)} The bandwidth $h$ of the Gaussian Kernel is selected via 5-fold cross-validation using the log-likelihood following \citet{pmlr-v80-ruff18a}. 

\paragraph{Semi-Supervised Anomaly Detection (SSAD)} A kernel method suggested by \citet{gornitz2013toward} which is a generalization to SVDD to both labeled and unlabled examples. Also granted the same unfair adavantage as OC-SVM/SVDD. 

\paragraph{Convolutional Autoencoder (CAE)} Autoencoders with convolution and deconvolution layers in the encoder and decoder, respectively. We use the same architectures described in \ref{net_arch} for our VAE.

\paragraph{Hybrid Methods}
In all of the the hybrid methods mentioned in the results, the inputs are representations from a converged autoencoder, instead of raw inputs.

\paragraph{Unsupervised Deep SVDD}
Two end-to-end variants of OC-SVM methods called Soft-Boundary Deep SVDD and One-Class Deep SVDD proposed by \citet{pmlr-v80-ruff18a}. They use an objective similar
to that of the classic SVDD to optimize the weights of a deep architecture.

\paragraph{Deep Semi-supervised Anomaly Detection (Deep SAD)} Recently proposed by \citet{ruff2019}, Deep SAD is a general method based on deep SVDD, which learns a neural-network mapping of the input that minimizes the volume of data around a predetermined point.

\paragraph{Semi-Supervised Deep Generative Models (SS-DGM)}
\citep{DBLP:journals/corr/KingmaRMW14} proposed a deep variational generative approach to semi-supervised learning. In this approach, a classifier is trained on the latent space embeddings of a VAE, which are lower in dimension than the original input. They also propose a probabilistic model that describes the data using the available labels. These two models are fused together to a stacked semi-supervised model.

\paragraph{Deep Structured Energy-Based Models (DSEBM)}
Proposed by \citet{DBLP:journals/corr/ZhaiCLZ16}, DSEBM is a deep neural technique, whose output is the energy function (negative log probability)
associated with an input sample. The chosen
architecture, as described by \citet{golan2018deep} is the same as that of the encoder part in the convolutional autoencoder used by OC-SVM Hybrid.

\paragraph{Deep Autoencoding Gaussian Mixture Model (DAGMM)} Proposed by \citet{zong2018deep}, DAGMM is an end-to-end deep neural network that leverages Gaussian Mixture Modeling to perform density estimation and unsupervised anomaly detection in a low-dimensional space learned by deep autoencoder.  It simultaneously
optimizes the parameters of the autoencoder and the mixture model in an end-to-end fashion, thus
leveraging a separate estimation network to facilitate the parameter learning of the mixture model.
The architecture, as described by \citet{golan2018deep} is the same as of the autoencoder we used is similar to that of the convolutional autoencoder used in OC-SVM Hybrid.

\paragraph{Anomaly Detection with a Generative Adversarial Network (ADGAN)}
A GAN-based model, proposed by \citet{deecke2018anomaly}. Anomaly detection is done with GANs by searching the generator's latent space for good sample representations. In the experiments performed by \citet{golan2018deep}, the generative model of the ADGAN had the same
architecture used by the authors of the original paper.

\paragraph{Deep Anomaly Detection using Geometric
Transformations (DADGT)} A stae-of-the-art deep anomaly detection in images method proposed by \citet{golan2018deep}. In this method, features are learned using a self-supervised paradigm -- by applying geometric transformations to the image and learning to classify which transformation was applied.

\paragraph{Inclusive Neural Random Fields (Inclusive-NRF)} 
A state-of-the-art energy-based model proposed by \citet{Song2018LearningNR}. The inclusive-NRF learns neural random fields for continuous data by developing inclusive-divergence minimized auxiliary generators and stochastic gradient sampling. As this model directly provides a density estimate, it is an efficient tool for AD, as the estimated density can be used as the decision criterion.

\section{Complete Ablative Analysis}
\label{apndx_ab_an}
The complete ablative analysis is reported in table \ref{tab:apndx_ablation_an}.

\begin{table}[ht]
%\vspace{-1.0em}
\begin{center}
\begin{tiny}
\begin{tabular}{llcccccccc}
        \toprule
                                               &                       &                   &   \textbf{Cardio} & \textbf{Satellite} & \textbf{CIFAR-10} \\
         \textbf{Encoder} & \textbf{Decoder}      & \textbf{Ensemble}   & \textbf{AUROC} & \textbf{AUROC} & \textbf{AUROC}  \\
        \midrule
         				Separate   & Unfreeze      & 5     & 96.6$\pm$0.6 & 65.3$\pm$1.1 & 49.8$\pm$10.7   \\
        	 				 Separate   &     Freeze              &  5 & 96.6$\pm$0.6  & 66.9$\pm$1.2 & 50.0$\pm$10.9  \\
        					 Same   &     Unfreeze              &  5 & 98.8$\pm$0.06 & 88.9$\pm$1.7  & 73.9$\pm$8.9 \\
        	 				 Same   &     Freeze              &  5 & \textbf{99.1$\pm$0.4} & \textbf{91.7$\pm$1.5} & \textbf{74.5$\pm$8.4}   \\
        	 				 Separate   & Unfreeze      & 1     & 82.5$\pm$3.3 & 63.1$\pm$5.3 & 50.4$\pm$10.2   \\
        	 				 Separate   &     Freeze              &  1 & 88.8$\pm$0.9  & 66.2$\pm$3.4 & 49.2$\pm$10.4  \\
        					 Same   &     Unfreeze              &  1 & 96.8$\pm$1.5 & 90.0$\pm$2.9  & 72.8$\pm$8.6 \\
        	 				 Same   &     Freeze              &  1 & 97.8$\pm$1.1 & 87.6$\pm$1.7 & 72.3$\pm$9.0  \\
        \bottomrule
        \end{tabular}
\end{tiny}
\end{center}
\caption{Complete ablative analysis of the Dual Prior method. AUROC is reported over average of 10 seeds for the \textit{satellite} and \textit{cardio}. For \textit{CIFAR-10}, results are reported for the experiment with 1\% of anomalies, averaged over 90 experiments.}
\label{tab:apndx_ablation_an}
\vspace{-0.5em}
\end{table}

\section{Implementation Details}\label{s:implementation_details}
We provide essential implementation information and describe how we tuned our models.
% \footnote{Our code will be published upon acceptance for publication}. 

\subsection{Network Architectures}
\label{net_arch}

Our work is based on deep variational autoencoders which require an encoder network and a decoder network. For the architectures, we follow \citet{ruff2019} architectures for the autoencoders, with the only differences being the use of bias weights in our architectures and instead of autoencoders we use \textit{variational} autoencoders (i.e. the encoder outputs mean and standard deviation of the latent variables).

For the image datasets, LeNet-type convolutional neural networks (CNNs) are used. Each convolutional layer is followed by batch normalization and leaky ReLU activations ($\alpha=0.1$) and $2\times2$-max pooling. On Fashion-MNIST, we use two convolutional layers, one the first with $16\times(5\times5)$ filters and second with $32\times(5\times5)$ filters. Following are two dense layers of 64 and 32 units respectively. On CIFAR-10, three convolutional layers of sizes $32\times(5\times5)$ ,$64\times(5\times5)$ and $128\times(5\times5)$ filters are followed by a final dense layer of 128 units (i.e. the latent space is of dimension 128). For MNIST, we use a different architecture where the first layer is comprised of $64\times(4\times4)$-filters and the second $128\times(4\times4)$-filter with ReLU activations and no batch normalization. We use two dense layers, the first with 1024 units and the final has 32 units. For CatsVsDogs we follow \cite{golan2018deep} architecture for the autoencoder. We employ three convolutional layers, each followed by batch normalization and ReLU activation. The number of filters in each layer are $128\times(3\times3)$, $256\times(3\times3)$ and $512\times(3\times3)$ respectively. The final dense layer has 256 units, which is the latent space dimension.

For the classic AD benchmark datasets, we use regular multi-layer perceptrons (MLPs). On arrhythmia, a 3-layer MLP with 128-64-32 units. On cardio, satellite, satimage-2 and shuttle, we use a 3-layer MLP with 32-16-8 units. On thyroid a 3-layer MLP with 32-16-4 units.

For the motion planning, we follow \citet{ichter_mp_18} and use two-layer MLP with 512 hidden units and latent space dimension of 256. We use dropout ($p=0.5$) as regularization and ReLU activation for the hidden layers. 

\subsection{Hyper-parameters and Tuning}
\label{hyper_param}
Tuning models in a semi-supervised setting is not a trivial task, as usually there is abundance of data from one class and a small pool of data from the other classes. Thus, it is not clear whether one should allocate a validation set out of the training set (and by doing that, reducing the number of available samples for training) or just evaluate the performance on the training set and hope for the best. \citet{ruff2019} didn't use a validation set, but predetermined the total number of epochs to run and finally evaluated the performance on the test set. We, on the other hand, decided to take a validation set out of the training set for the image datasets, as we have enough data. The validation set is composed of unseen samples from the normal class and samples from the \textit{current} outlier class unlike the test set, which is composed of samples from all ten classes (9 outlier classes). On the classic AD benchmark datasets, as there is very few outlier data, we evaluate the performance during training on the training set itself without taking a validation set. Finally, we evaluate the performance on the test set. For all datasets, we used a batch size of 128 and an ensemble of 5 VAEs. For the image datasets, we run 200 epochs and for the classical AD benchmarks we run 150 epochs. For the MML method, we found $\gamma=1$ to perform well on all datasets. Furthermore, similarly to the additional hyper-parameter $\beta_{KL}$ in the ELBO term, we add $\beta_{CUBO}$ to the CUBO term as derived in Equation \ref{eq:full_cubo_eq}. The motivation for adding this balancing hyper-parameter is that since $ \mathcal{L} = \exp\{n \cdot CUBO_n(q_\phi(z|x))\}$ reaches the same optima as $CUBO_n(q_\phi(z|x))$ as explained earlier, so does $ \mathcal{L} = \exp\{\beta \cdot n \cdot CUBO_n(q_\phi(z|x))\}$. Finding a good value for $\beta_{CUBO}$ can contribute to the performance, as reflected in the results. Moreover, for the optimization of the CUBO component, we used gradient clipping and learning rate scheduling (every 50 epochs, learning rate is multiplied by 0.1). Tables \ref{tab:dp_hyp} and \ref{tab:cml_hyp} summarize all the hyper-parameters chosen for our model per dataset.

\begin{table}[ht]
%\vspace{-1.0em}
\begin{center}
\begin{scriptsize}
\begin{tabular}{llcccccccc}
        \toprule
        \textbf{Data}	    & \textbf{ND Update Interval} & \textbf{Learning Rate} & \textbf{$\beta_{KL}$}   & \textbf{$\alpha$} \\
        \midrule
        MNIST                     & 2   & 0.001      & 0.005     & 10  \\
        Fashion-MNIST	 				& 2   &     0.001 &  0.005 & 10  \\
        CIFAR-10					& 2   &     0.001             &  0.005 & 10   \\
        CatsVsDogs	 				& 1   &     0.001              &  0.005 & 10   \\
         \midrule
        arrhythmia    				& 1   & 0.0005      & 0.5     & 2 \\
        cardio	 				& 1   &     0.001              &  0.05 & 5   \\
        satellite					& 1   &     0.001              &  0.05 & 5   \\
        satimage-2	 				& 1   &     0.0005              &  0.05 & 10   \\
        shuttle	 				& 1   &     0.001              &  0.05 & 5   \\
        thyroid	 				& 1   &     0.0001              &  0.05 & 10  \\
        \bottomrule
        \end{tabular}
\end{scriptsize}
\end{center}
\caption{Hyper-parameters for the Dual Prior VAE}
\label{tab:dp_hyp}
\vspace{-0.5em}
\end{table}

\begin{table}[ht]
%\vspace{-1.0em}
\begin{center}
\begin{scriptsize}
\begin{tabular}{llcccccccc}
        \toprule
        \textbf{Data}	    & \textbf{ND Update Interval} & \textbf{Learning Rate} & \textbf{$\beta_{KL}$}   & \textbf{$\beta_{CUBO}$} \\
        \midrule
        MNIST                     & 2   & 0.0005 (0.0002)     & 0.005     & 0.005 (0.02)  \\
        Fashion-MNIST	 				& 2   &     0.001 &  0.005 & 0.005 (0.25)  \\
        CIFAR-10					& 2   &     0.001            &  0.005 & 0.005   \\
        CatsVsDogs	 				& 1   &     0.0005              &  0.005 & 0.005   \\
         \midrule
        arrhythmia    				& 1   & 0.0005      & 0.5     & 0.5 \\
        cardio	 				& 1   &     0.001              &  0.05 & 0.05   \\
        satellite					& 1   &     0.001              &  0.05 & 0.05   \\
        satimage-2	 				& 1   &     0.001              &  0.05 & 0.05   \\
        shuttle	 				& 1   &     0.001              &  0.05 & 0.05   \\
        thyroid	 				& 1   &     0.0001              &  0.05 & 0.05  \\
        \bottomrule
        \end{tabular}
\end{scriptsize}
\end{center}
\caption{Hyper-parameters for the Max-Min Likelihood VAE. In parenthesis, changes for the setting with polluted samples.}
\label{tab:cml_hyp}
\vspace{-0.5em}
\end{table}

For training, we follow \citet{2016arXiv160202282K} recommendations for training VAEs and use a 20-epoch annealing for the KL-divergence component, that is, the KL-divergence coefficient, $\beta_{KL}$, is linearly increased from zero to its final value. Moreover, we allow the VAE to first learn a good representation of the normal data in a warm-up period of 50 epochs, and then we begin applying the novelty detection updates. For optimization, we use Adam \citep{adam_14} with a learning rate schedule to stabilize training when outlier samples are fed after the warm-up period.

\section{CUBO Loss Derivation}
\label{cubo_deriv}
The $\chi$-divergence is defined as follows: $$ D_{\chi^2}(p \Vert q) = \E_{q(z;\theta)}\big[\big(\frac{p(z | x)}{q(z;\theta)} \big)^2 - 1 \big] $$
The general upper bound as derived in \cite{dieng2017variational}: $$ \mathcal{L}_{\chi^2}(\theta) = CUBO_2 = \frac{1}{2}\log\E_{q(z;\theta)}\big[\big(\frac{p(x,z)}{q(z;\theta)} \big)^2\big] $$
The optimized CUBO: $$ \mathcal{L} = \exp\{2 \cdot CUBO_2(\theta)\} =  \E_{q(z;\theta)}\big[\big(\frac{p(x,z)}{q(z;\theta)} \big)^2\big]$$
In the our VAE framework, we denote $\mathcal{L}_R = -\log p(X|z)$, the reconstruction error between the output of the decoder and the original input. We assume that $q(z|X) \sim \mathcal{N}(\mu_q(X), \Sigma_q(X))$ and that in general, $p_{outlier}(z) \sim \mathcal{N}(\mu_o, I)$, but we note that in our case, we set $\mu_o=0$, for both CUBO and hybrid methods. As the CUBO loss is derived only for the anomalous data, we omit the labels in the following. $\beta_{CUBO}$ is a balancing hyper-parameter we add, as mentioned in \ref{hyper_param}.
$$ \mathcal{L}_{CUBO} =  \E_{q(z;\theta)}\big[\big(\frac{p(x,z)}{q(z;\theta)} \big)^2\big] = \E_{q}\big[ \exp\{2\log p(X|z) +2\log\big( \frac{p(z)}{q(z|X)} \big) \} \big] = $$  $$  \E_{q}\big[ \exp\{-2\cdot \mathcal{L}_R +2\beta_{CUBO}\cdot(\log p(z) -\log q(z|X) ) \} \big] =  $$ $$ \E_{q}\big[ \exp\{-2\cdot \mathcal{L}_R +2\beta_{CUBO}\cdot\big( -\frac{1}{2}(z - \mu_o)^T(z - \mu_o) - [-\frac{1}{2}\log |\Sigma_q| -\frac{1}{2}(z-\mu_q)^T\Sigma_q^{-1}(z -\mu_q)] \big) \} \big] =  $$ \begin{multline*} \E_{q}\big[ \exp\{-2\cdot \mathcal{L}_R +\beta_{CUBO}\cdot\big( -[z^Tz-2z^T\mu_o +\mu_o^T\mu_o] + \log |\Sigma_q| + \\ z^T\Sigma_q^{-1}z -2z^T\Sigma_q^{-1}\mu_q +\mu_q^T\Sigma_q^{-1}\mu_q \big) \} \big] = \end{multline*} \begin{multline*} \E_{q}\big[ \exp\{-2\cdot \mathcal{L}_R +\beta_{CUBO}\cdot\big(  \log |\Sigma_q| +\mu_q^T\Sigma_q^{-1}\mu_q - \mu_o^T\mu_o -  \\  z^Tz + 2z^T\mu_o + z^T\Sigma_q^{-1}z -2z^T\Sigma_q^{-1}\mu_q  \big) \} \big] = \end{multline*} 
\begin{multline*}  \exp\{\beta_{CUBO}\cdot\big(  \log |\Sigma_q| +\mu_q^T\Sigma_q^{-1}\mu_q - \mu_o^T\mu_o \big)\} \cdot \\ \E_{q}\big[\exp\{-2\cdot \mathcal{L}_R + \beta_{CUBO} \cdot \big( -  z^Tz + 2z^T\mu_o + z^T\Sigma_q^{-1}z -2z^T\Sigma_q^{-1}\mu_q  \big) \} \big] = \end{multline*}
\begin{multline*}  \exp\{\beta_{CUBO}\cdot\big(  \log |\Sigma_q| +\mu_q^T\Sigma_q^{-1}\mu_q - \mu_o^T\mu_o \big)\} \cdot \\\exp\{\log \E_{q}\big[\exp\{-2\cdot \mathcal{L}_R +\beta_{CUBO} \cdot \big( -  z^Tz + 2z^T\mu_o + z^T\Sigma_q^{-1}z -2z^T\Sigma_q^{-1}\mu_q  \big) \} \big]\} = \end{multline*}
\begin{multline} \label{eq:full_cubo_eq}  \exp\{\beta_{CUBO}\cdot\big(  \log |\Sigma_q| +\mu_q^T\Sigma_q^{-1}\mu_q - \mu_o^T\mu_o \big) + \\ \log \E_{q}\big[\exp\{-2\cdot \mathcal{L}_R +\beta_{CUBO} \cdot \big( -  z^Tz + 2z^T\mu_o + z^T\Sigma_q^{-1}z -2z^T\Sigma_q^{-1}\mu_q  \big) \} \big] \} \end{multline}
The expectation is estimated with Monte Carlo and for numeric stability we employ the commonly used \textit{log-sum-exp trick}.

\section{ELBO With Gaussian Prior}
We provide the derivation for the ELBO objective function where the prior, $p(z)$ is a Gaussian with non-zero mean, that is, $z \sim \mathcal{N}(\mu_o, I)$.
Under the i.i.d. assumption in the VAE framework, we assume that $z_i \sim \mathcal{N}(\mu_{o_i}, 1)$ and $ z_i | x \sim \mathcal{N}(\mu_{q_i}, \sigma_{ii}^2)$. Thus, it holds that: $$ \E_q[z_i^2] = \sigma_{ii}^2 + \mu_{q_i}^2 $$
We now derive the KL-divergence component of the ELBO: $$ \KL[q(z|X) \Vert p(z)]= \E_q\big[ \frac{\log q(z|X)}{\log p(z)} \big]=\E_q[\log q(z|X)] - \E_q[\log p(z)] $$
$$\E_q[\log q(z|X)] = -\frac{1}{2} \log |\Sigma_q| -\frac{1}{2}\E_q[(z-\mu_q)^T\Sigma_q^{-1}(z-\mu_q)] =  $$
$$     -\frac{1}{2} \log |\Sigma_q| + \frac{1}{2} \mu_q^T\Sigma_q^{-1}\mu_q -\frac{1}{2}\E_q[z^T\Sigma_q^{-1}z] = $$ $$ -\frac{1}{2}\sum_{i=1}^n [\log \sigma_{ii}^2 -\frac{\mu_{q_i}^2}{\sigma_{ii}^2} + 1 + \frac{\mu_{q_i}^2}{\sigma_{ii}^2}]= -\frac{1}{2}\sum_{i=1}^n [\log \sigma_{ii}^2 + 1] $$
$$ \E_q[\log p(z)] = -\frac{1}{2} \E_q[(z - \mu_o)^T(z-\mu_o)] =  $$ $$ \frac{1}{2} \sum_{i=1}^n [\sigma_{ii}^2 +\mu_{q_i}^2 -2\mu_{q_i}\mu_{o_i} + \mu_{o_i}^2] $$
Finally: $$ \KL[q(z|X) \Vert p(z)]= -\frac{1}{2}\sum_{i=1}^n[1 + \log \sigma_{ii}^2 -\sigma_{ii}^2 -\mu_{q_i}^2 +2\mu_{q_i}\mu_{o_i} - \mu_{o_i}^2] $$

\section{Deep SVDD/SAD with Ill-trained Autoencoder}
\label{apndx_deepsad10}
To demonstrate that Deep SAD and Deep SVDD are highly dependent on the representation ability of the autoencoder, we train the autoencoder for 10 epochs instead of 150, and then train normally using the objective of Deep SAD/Deep SVDD. Experiment was conducted on the CIFAR-10 dataset. The results presented in Table \ref{tab:deepsad10} clearly show Deep SVDD/SAD's objective function has no effect on anomaly detection.

 \begin{table}[ht]
%\vspace{-1.0em}
\begin{center}
\begin{normalsize}
\begin{tabular}{llcccccccc}
        \toprule
                                          \textbf{Anomaly}    &          \textbf{CIFAR-10}    \\
         \textbf{Ratio} & \textbf{AUROC}        \\
        \midrule
         				0.0   &  59.8$\pm$9.8    \\
        	 				 0.01   &      59.4$\pm$9.6  \\
        					 0.05   &     58.8$\pm$9.4  \\
        	 				 0.1   &      58.4$\pm$10.0       \\
        	 				 0.2   &      56.9$\pm$11.2      \\
        \bottomrule
        \end{tabular}
\end{normalsize}
\end{center}
\caption{Deep SAD results for 10 epochs of autoencoder pre-training on \textit{CIFAR-10}. Results are reported for the experiment are averaged over 90 experiments. For no anomalies, Deep SAD reduces to Deep SVDD.}
\label{tab:deepsad10}
\vspace{-0.5em}
\end{table}

\section{Sentiment Analysis Experiment Details}
\label{apndx_nlp}
In this section, we further demonstrate the generality of our method by considering a Natural Language Processing (NLP) application.
One of the most important tasks in NLP is the task of \textit{sentiment analysis}. The objective is to identify opinions expressed in a piece of text, e.g., to determine whether the writer's attitude towards a particular topic is positive, negative, or neutral. \citet{bowman-etal-2016-generating} proposed learning latent representations of entire sentences using a recurrent neural network (RNN) based VAE. Typically, sentiment analysis is learned in a supervised setting. Here, we consider identifying sentiment anomalies. E.g., having seen only data for positive opinions, identify when a negative opinion is encountered. Using our SSAD method, we shall show that even a small number of labelled outliers can significantly improve detection.

% We adapt this VAE for the task of sentiment analysis on the 
We experiment on the IMDB review dataset, which contains 30K positive and negative movie reviews. 
In our experiment, we use only the first sentence of the review (which leads to a noisy dataset as the first sentence does not necessarily express the reviewer opinion, e.g., includes a short summary of the movie). Unlike images, text data does not have a fixed size or structure. Moreover, samples from different domains/sentiments can have the same structure (e.g. "this is the best movie"--"this is the worst movie"), with the difference being limited only to certain words. Since the VAE represents whole sentences in the latent space, it seems that detecting sentiment novelties would be challenging. 

% \subsection{Pre-processing and Implementation}
We first take the first sentence of each review that has at least 5 words and no more than 20 words. We then separate to train (0.7) and test (0.3) sets and build a vocabulary of word representations using GloVe \citep{Pennington14glove:global}. We follow the same architecture and training procedure as in the original paper \citep{bowman-etal-2016-generating}, including annelaing for the KL term and word dropout (0.5), and train with unlabeled data for 480 epochs. Finally, we supply labeled anomalies for 5 epochs and measure the AUROC. As for the MML-VAE parameters, we used $\gamma=1$.

We demonstrate results using MML-VAE as the SSAD method in Table \ref{tab:apndx_imdb}.As our results show, using only 1\% of outliers (i.e., reviews from the opposing sentiment), leads to a significant improvement.

\begin{table}[ht]
%\vspace{-1.0em}
\begin{center}
\begin{normalsize}
\begin{tabular}{llcccccccc}
        \toprule
                                 \textbf{Outlier}              &       \textbf{Positive}                &    \textbf{Negative} & \textbf{Mean}    \\
         \textbf{Ratio} & \textbf{AUROC}      & \textbf{AUROC}  & \textbf{AUROC}    \\
        \midrule
         				0.0   & 52.8$\pm$0.01      & 60.9$\pm$0.004 & 56.9$\pm$4.0     \\
        	 		    0.01   &     56.6$\pm$5.0  &  64.8$\pm$1.5  & 60.7$\pm$5.4  \\
        \bottomrule
        \end{tabular}
\end{normalsize}
\end{center}
\caption{Complete MML-VAE AUROC results for the task of sentiment analysis on the IMDB dataset. Each column corresponds to the unlabeled train data. The results are averaged over 5 runs.}
\label{tab:apndx_imdb}
\vspace{-0.5em}
\end{table}

\end{document}